\documentclass[runningheads]{llncs}

\usepackage{eccv}

\usepackage{eccvabbrv}

\usepackage{graphicx}
\usepackage{booktabs}
\usepackage{booktabs}
\usepackage[table]{xcolor}
\usepackage{multirow}
\usepackage{subcaption}
\usepackage{wrapfig}

\makeatletter
\renewcommand\section{\@startsection{section}{1}{\z@}%
                       {-9\p@ \@plus -2\p@ \@minus -2\p@}%
                       {6\p@ \@plus 2\p@ \@minus 2\p@}%
                       {\normalfont\large\bfseries\boldmath
                        \rightskip=\z@ \@plus 8em\pretolerance=10000 }}
\renewcommand\subsection{\@startsection{subsection}{2}{\z@}%
                       {-9\p@ \@plus -2\p@ \@minus -2\p@}%
                       {4\p@ \@plus 2\p@ \@minus 2\p@}%
                       {\normalfont\normalsize\bfseries\boldmath
                        \rightskip=\z@ \@plus 8em\pretolerance=10000 }}
\makeatother
\usepackage[accsupp]{axessibility}  

\usepackage{hyperref}

\usepackage{orcidlink}

\def\ours{MM-Nav\xspace}

\def\ours{MM-Nav\xspace}

\definecolor{color1}{HTML}{f29a6c}
\definecolor{color2}{HTML}{5eba94}

\definecolor{mygray}{HTML}{f0f0f0}

\definecolor{reaching}{HTML}{2F5597}
\definecolor{squeezing}{HTML}{548235}
\definecolor{avoiding}{HTML}{C55A11}

\begin{document}

\title{MM-Nav: Multi-View VLA Model for Robust Visual Navigation via Multi-Expert Learning} 

\titlerunning{MM-Nav}

\newcommand{\equalcontrib}{\textsuperscript{*}}
\newcommand{\corrauth}{\textsuperscript{\ensuremath{\dagger}}}

\newcommand{\titlefootnote}[1]{%
  \begingroup
  \renewcommand{\thefootnote}{}%
  \footnotetext[0]{#1}%
  \endgroup
}

\author{Tianyu Xu\inst{1,2}\equalcontrib\orcidlink{0009-0003-5653-9902} \and
Jiawei Chen\inst{1}\equalcontrib\orcidlink{0009-0001-2821-3484} \and
Jiazhao Zhang\inst{1,2}\equalcontrib\orcidlink{0000-0001-9459-293X} \and
Wenyao Zhang\inst{2,3}\orcidlink{0009-0006-3090-255X} \and
Zekun Qi\inst{2,4}\orcidlink{0009-0001-2554-5141} \and
Minghan Li\inst{2}\orcidlink{0009-0009-5153-2406} \and
Jiahang Liu\inst{1,2}\orcidlink{0009-0002-3334-6692} \and
Lu Yue\inst{1,2}\orcidlink{0009-0006-8946-2704} \and
Zhizheng Zhang\inst{2}\corrauth\orcidlink{0000-0002-5360-7565} \and
He Wang\inst{1,2}\corrauth\orcidlink{0000-0002-3365-4620}}


\authorrunning{T.~Xu et al.}


\institute{
Peking University, Beijing, China \and
Galbot, Beijing, China \and
Shanghai Jiao Tong University, Shanghai, China \and
Tsinghua University, Beijing, China
}

\maketitle

\titlefootnote{\textsuperscript{*}: Equal contributions, \textsuperscript{\ensuremath{\dagger}}: Corresponding authors. \\Project page: \url{https://pku-epic.github.io/MM-Nav-Web/}}


\begin{figure}[t]
  \centering
  \vspace{-0.5em}
  \includegraphics[width=\linewidth]{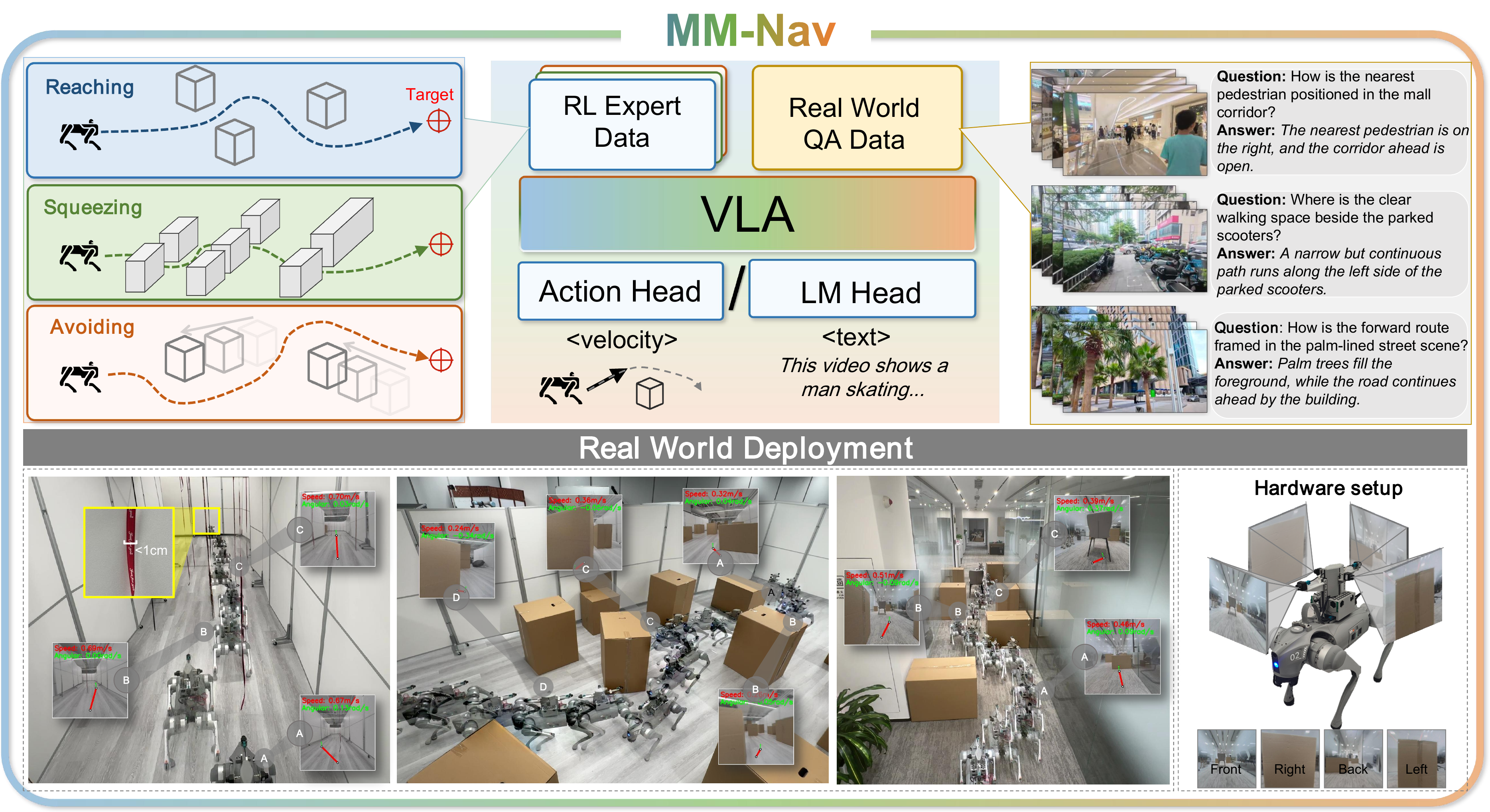}

  \vspace{-0.3em}
  \noindent
  \begin{minipage}[t]{0.75\linewidth}
    \centering

  \end{minipage}%
  \hfill
  \begin{minipage}[t]{0.2\linewidth}
    \centering
    
  \end{minipage}

  \vspace{-0.3em}
  \caption{MM-Nav is a multi-view VLA policy trained from capability-specific RL expert trajectories together with large-scale QA data, and outputs continuous velocity commands via an action head.
\textbf{Bottom-left}: real-world rollouts illustrating robust navigation in challenging scenes, including thin-obstacle (wire) avoidance and squeezing through cluttered/transparent structures.
\textbf{Bottom-right}: robot platform and the 360° surround-view camera configuration used for deployment.}
  \label{fig:teaser}
  \vspace{-0.8em}
\end{figure}

\begin{abstract}
    Visual navigation policies are widely regarded as a critical research direction, as they emulate human navigation behavior by leveraging egocentric visual observations.
    However, unlike LiDAR point clouds or depth maps, visual observations do not provide explicit geometric information for navigation, especially in cluttered or dynamic environments, which motivates the need for learning-based models and large-scale data. 
    To this end, we propose to leverage Vision-Language-Action (VLA) models to learn diverse navigation capabilities from synthetic expert data and to alleviate the sim-to-real gap by co-training on large-scale real-world Visual Question Answering (VQA) data. 
    Specifically, we develop \ours, a $7\text{B}$ multi-view VLA model featuring custom-designed architectures and enabling a 7 Hz inference speed with $360^{\circ}$ observation.
    For large-scale navigation data, 
    we collect a total of 1.5 million expert demonstrations from three reinforcement learning (RL) experts, each trained with privileged information in a challenging, tailor-made environment and specialized in one of three navigation capabilities: \textit{reaching}, \textit{squeezing}, and \textit{avoiding}.
    We then iteratively train \ours on these data, dynamically balancing the training data ratio across the three capabilities based on their respective performance.
    Through extensive experiments in synthetic and real-world environments, we demonstrate that our model achieves strong performance and generalization on different benchmarks. \ours obtains a success rate of 88.1\% on the InternVLA-N1 System-1 point-goal navigation benchmark. Moreover, we find that our student VLA model outperforms the RL teachers, demonstrating the synergistic effect of integrating multiple capabilities. Extensive real-world experiments further confirm the effectiveness of our method. 
    %
\end{abstract}


\section{Introduction}
\label{sec:intro}

Visual navigation has garnered considerable attention in the robotics field~\cite{shah2023gnm,shah2023vint,sridhar2024nomad,hirose2019deep,meng2020scaling,kim2025enhancing,zhang2024uni,zhang2024navid,wang2025trackvla,Xu_2025_ICCV,Zhong_2025_ICCV,Wang_2025_ICCV}, as it requires robots to reach target locations based on visual inputs. Visual observations provide detailed and rich environmental information for navigation while remaining cost-effective. However, in complex and cluttered navigation environments, interpreting such informative visual data and planning appropriate navigation actions remains challenging~\cite{sridhar2024nomad,shah2023vint,Wang_2025_ICCV}, demanding highly intelligent models and large-scale navigation datasets.

To this end, existing methods~\cite{shah2023gnm,zhang2024uni,wang2025trackvla,shah2023vint,sridhar2024nomad} learn visual navigation policies from synthetic data or teleoperated demonstrations, with recent works further leveraging real-world passive navigation videos to emulate general navigation behaviors~\cite{cheng2025navila,pmlr-v270-hirose25b,liu2025citywalker,hirose2025driveanywhere}.However, real-world data collection is expensive and often restricted to limited camera and scene configurations~\cite{chhablani2025embodiedsplat,pan2025lookout} (\textit{e.g.}, front-view only).
While synthetic data can be generated more flexible and efficiently, it typically suffers from significant sim-to-real gaps due to non-photorealistic rendering~\cite{cai2025navdp,wang2024grutopia,meng2025aim}. 
Moreover, both paradigms are largely confined to relatively spacious environments~\cite{cai2025navdp,eftekhar2024one} and rarely include highly challenging or hazardous scenarios. This is because collecting real-world data for collision avoidance is costly and risky~\cite{lee2026rvnbench,roth2025fdm}, while generating such data in simulation through rule-based approaches remains nontrivial~\cite{zhu2026hicrowd,seneviratne2025halo,Qin_2025_ICCV}.

To address these limitations, we propose \ours, a multi-view Vision-Language-Action (VLA) model that captures 360$^\circ$ observations around the robot, learns navigation capabilities from multiple synthetic RL experts, and mitigates the sim-to-real gap by co-training on large-scale real-world Visual Question Answering (VQA) data, as shown in \Cref{fig:teaser}.
To overcome the lack of highly challenging scenarios in existing navigation datasets, we construct synthetic environments and train separate RL experts for three distinct skills: reaching, squeezing, and avoiding, which are difficult and risky to obtain in the real world.
We collect 500k successful expert demonstrations to form a capability-diverse training dataset that is challenging to obtain in real-world settings.
Specifically, we first initialize \ours on the collected expert demonstrations and then iteratively refine it with an online teacher-student strategy in a DAgger manner~\cite{ross2011reduction}. Beyond standard DAgger, we dynamically balance the training data ratio across the three capabilities based on their respective performance, which facilitates stable improvement and accelerates convergence.
Moreover, building upon a VLM framework~\cite{zhang2025navfom}, we co-train \ours on 300k real-world VQA data. This enables the model to leverage the real-world visual understanding from VQA, thereby strengthening its generalization ability and mitigating the sim-to-real gap.

We conduct extensive experiments in both synthetic and real-world environments. The results demonstrate that \ours achieves strong navigation performance across diverse capability settings, even outperforming specifically trained RL experts. Notably, on the InternVLA-N1 System-1 point-goal navigation benchmark~\cite{wang2025internvla}, \ours achieves an average SR of 88.1\%, outperforming the best previous method (72.2\%) by +15.9 points. Extensive real-world evaluations further confirm its robust zero-shot sim-to-real transfer ability in challenging environments. In addition, our efficient tokenization design enables inference at approximately 7 Hz, comparable to existing visual navigation methods~\cite{zhang2024uni,shah2023vint}.

The primary contributions can be summarized as follows:
(1) \textit{Navigation data}: we build tailored multi-capability simulation environments and train privileged RL experts to provide diverse supervision for reaching, squeezing, and dynamic obstacle avoidance.
(2) \textit{Training strategy}: we introduce a capability-balanced iterative refinement procedure that mitigates policy-distribution shift and alleviates capability imbalance across different navigation skills.
(3) \textit{Model design}: we develop an efficient 7B multi-view VLA policy that uses $360^\circ$ RGB observations and directly predicts continuous omnidirectional velocity at 7 Hz.

\section{Related Works}
\noindent\textbf{Learning-based small navigation model.}
A considerable amount of learning-based navigation relies on models with a smaller parameter count. These models are often computationally efficient for specific tasks but face limitations in generalization and adaptability to complex scenarios. They can be broadly categorized into imitation learning and reinforcement learning. 
Imitation Learning (IL) acquires navigation capabilities by learning from expert demonstrations. This approach is relatively data-efficient as the expert data inherently contains effective strategies~\cite{cai2025navdp,sridhar2024nomad,shah2023vint,kim2025enhancing,xiao2025anycar,hirose2019deep,hirose2023sacson,ehsani2024spoc,cai2024bridging,eftekhar2024one,lu2024pret,xu2024disco,cui2024frontier}. For instance, NavDP~\cite{cai2025navdp} and LoGoPlanner~\cite{peng2025logoplanner} utilizes an A* planner with access to privileged information to generate high-quality expert trajectories.
Reinforcement Learning (RL) offers a structured approach to end-to-end navigation by leveraging modern simulation technologies~\cite{he2024agile,xu2025navrl,wang2025omni,yao2025towards,zeng2024poliformer,he2026seeing,Qin_2025_ICCV}. These methods are often limited to model size and face strong difficulties in handling the sim-to-real gap.
In contrast, our method leverages VLA, utilizing VLA's inherent generalization capabilities to address the sim-to-real transfer challenge. 
    
\noindent\textbf{Learning-based large navigation model.}
Large model learning~\cite{chiang2023vicuna,liu2023llava,zhu2023chatgpt,kim2024openvla, zhang2025dreamvla, sun2025view,hirose2025omnivla,zhang2025navfom,castro2025vamos,yang2025lohovla,xue2026omninav,wang2026imaginenavpp,zhang2025mem2ego,wang2025internvla}, particularly with Vision-Language-Action models, has significantly advanced navigation by leveraging powerful generalization and semantic understanding capabilities~\cite{zhang2024uni,zhang2024navid,wang2025trackvla,yokoyama2024vlfm,cheng2025navila,castro2025vamos,huang2026ticvla,yu2025correctnav,shi2025fastsmartway,qiao2024llm}. An approach employs off-the-shelf large models in a zero-shot manner~\cite{long2024instructnav,long2024discuss,shah2022robotic,shah2023lm}, converting visual scenes into text descriptions for a Large Language Model to perform high-level planning~\cite{zhou2023navgpt,zhou2025navgpt,song2023llm,huang2022visual,qi2025sofar}. However, this vision-to-text abstraction creates an information bottleneck, often limiting the agent to sparse landmarks and static environments. End-to-end VLA models like NaVid\cite{zhang2024navid} and Uni-NaVid\cite{zhang2024uni} tackle tasks ranging from object-search navigation and human-following to open-ended instruction following, all through a unified language-guided policy.  In contrast to these models, which are constrained by a discrete action space (e.g., FORWARD, TURN LEFT)~\cite{zhang2024uni,zhang2024navid}, our VLA policy directly outputs continuous velocity commands, enabling the agile and instantaneous responses required for robust real-world deployment.

\noindent\textbf{Visual-only navigation.}
Visual-only navigation relies solely on RGB images to guide movement, which is more economical and lightweight for practical deployment.  A primary challenge for such methods is the implicit nature of 3D geometry in 2D images, as pure RGB input lacks direct depth information, making robust obstacle avoidance non-trivial~\cite{yang2024depth}.
Foundational models like GNM~\cite{shah2023gnm}, ViNT~\cite{shah2023vint}, Scaling~\cite{meng2020scaling}, and NoMaD~\cite{sridhar2024nomad} have demonstrated impressive generalization through imitation learning on large-scale datasets, but their reliance on single-camera views limits their spatial awareness. To mitigate safety risks, CARE ~\cite{kim2025enhancing} estimates RGB depth to add a collision avoidance layer on top of a pre-trained policy.
Our work addresses these limitations using a four-camera surround-view system for 360° perception~\cite{hirose2019deep}. To tackle the challenge of learning from depthless RGB data, we leverage the comprehensive ability of VLA. We adopt the policy distillation paradigm, proven effective in works like X-Nav~\cite{wang2025x} and COMPASS~\cite{liu2025compass}. But with a key innovation: we train multiple RL experts on specific capabilities using privileged depth information. This expertise is then distilled into a unified VLA policy that robustly navigates by inferring environmental geometry from surround-view RGB images only.

\section{Method}
\begin{figure*}[t]
    \centering
    \includegraphics[width=\linewidth]{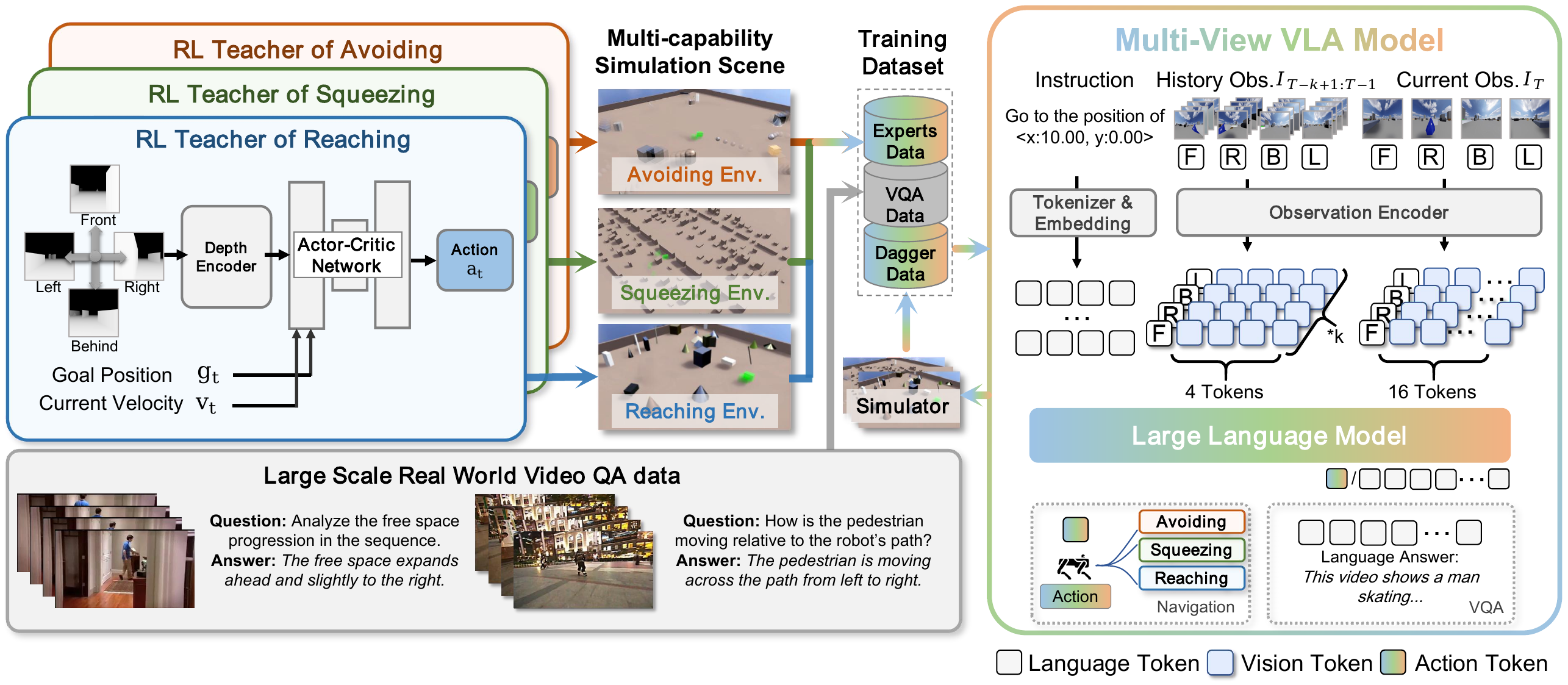}
    \caption{\textbf{Pipeline of \ours}. Our proposed teachers-student training pipeline. Independent RL teachers are trained in different scenes for multi-capability and distill knowledge to the VLA student. Further, the student is deployed in the capability-specific simulation scene to be iteratively fine-tuned.}
    
    \vspace{-1.5mm}
    \label{fig:pipeline}
\end{figure*}

\subsection{Overview}

\noindent\textbf{Task Definition.}  
We formulate the task as learning a velocity control policy \(\pi\) for an omnidirectional robot to navigate safely to a specific point goal in a cluttered and dynamic environment. At each time step \(t\), given the position of a point goal $g_t=[g_t^\text{x},g_t^\text{y}]$ and online captured multi-view RGB frames $O_t=\{I_{t-k+1:t}^N\}$ where $I_t^N \in \mathbb{R}^{3\times H \times W}$ ($N$ denotes the number of cameras, including 4 camera views $\{I^\text{front}, I^\text{right}, I^\text{back}, I^\text{left}\}$), the policy $\pi(O_t, g_t) \mapsto {a}_t$ predicts an action $a_t = [v_t^\text{x}, v_t^\text{y}, v_t^\text{yaw}]$, representing omnidirectional velocity. 
We adopt velocity control because it allows the robot to execute arbitrary omnidirectional motions on a two-dimensional plane, thus enhancing its responsiveness in obstacle avoidance~\cite{liu2025compass,xu2025navrl,yao2025towards}. The objective is to ensure the velocities generated by the policy \(\pi\) are collision-free and reach the designated goal.

\noindent\textbf{Pipeline Overview.}  
Our pipeline comprises two steps: (1) training of multiple RL experts with different capabilities and initial VLA finetuning; (2) teachers-student online training iteration between RL experts and VLA.
As shown in \Cref{fig:pipeline}, we first use reinforcement learning to train three capability-specific experts in simulation, including reaching, squeezing, and avoiding. These capabilities are obtained specifically by training privileged RL experts in tailored scenarios. Second, different capability navigation trajectories from RL experts are generated and collected, which are then used to directly train a VLA model to initialize a general navigation policy mastering different capabilities. Finally, the basic VLA model (student) is deployed in the simulation environment, and we online collect the expert action of the RL teachers for further finetuning the VLA model. This mechanism is conducted iteratively until performance converges.

\subsection{RL Experts for Different Navigation Capabilities}

\begin{figure*}[t]
    \centering
    \includegraphics[width=\linewidth]{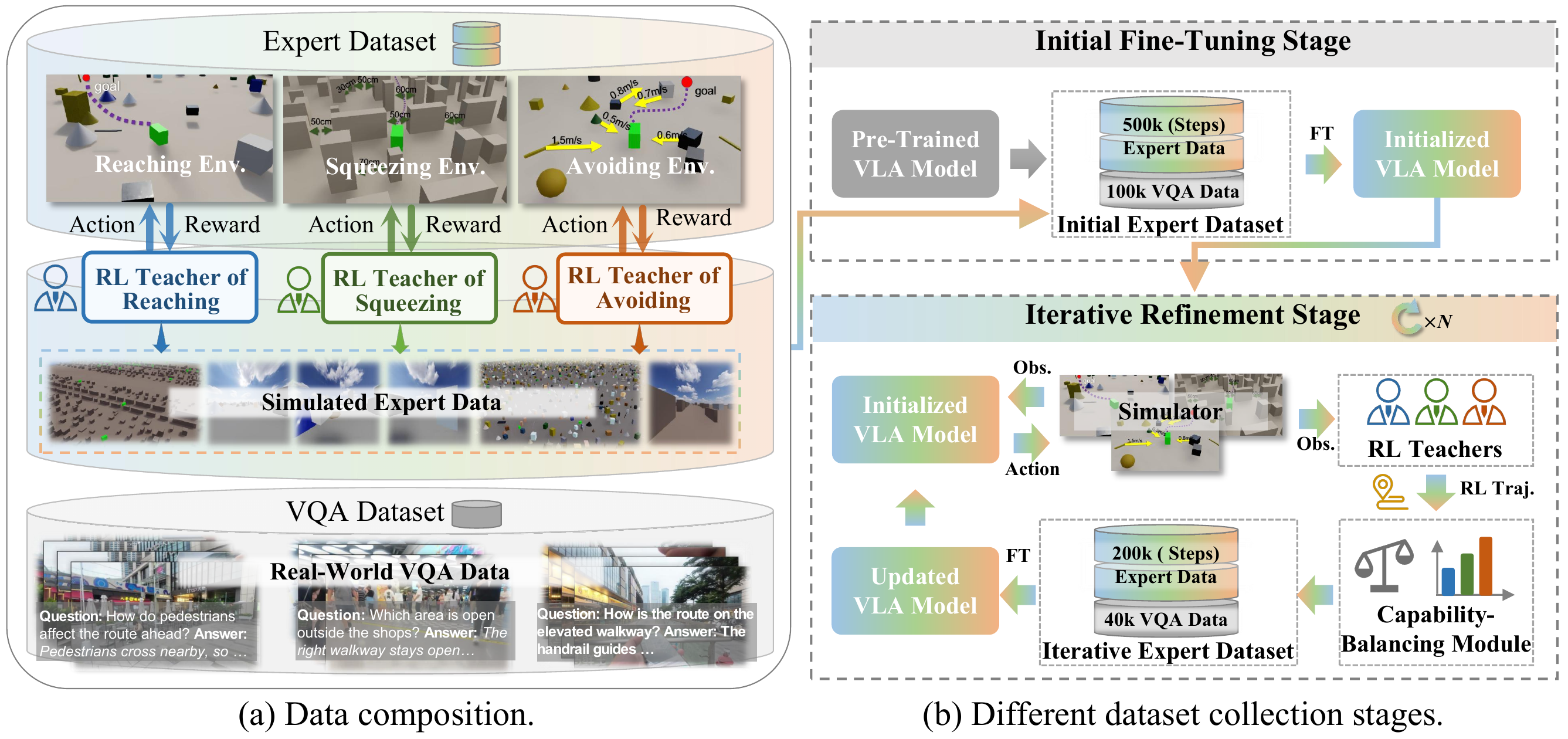}
    \caption{\textbf{Training data composition and collection stages of MM-Nav.}
(a) Dataset composition. MM-Nav is trained on simulated expert trajectories from three RL teachers (reaching, squeezing, avoiding) together with real-world VQA data.
(b) Dataset collection stages. The VLA model is initialized with successful teacher demonstrations and real-world VQA data, then iteratively deployed in three environments to collect teacher-supervised trajectories, with a capability-balancing module adjusting data ratios for fine-tuning.}
    \vspace{-1.5mm}
    
    \label{fig:strategy}
\end{figure*}

We construct three distinguishing environments for training RL experts to obtain different navigation capabilities: (1) reaching: approach and reach a specific point goal while avoiding static obstacles, (2) squeezing: squeeze through cluttered and narrow gaps between obstacles and walls, and (3) avoiding: actively avoid crowded dynamic obstacles moving at random speed, as illustrated in \Cref{fig:strategy}.


\noindent\textbf{Reaching environments.}  
We construct the reaching scene which contains randomized static obstacles, generated in different shapes (cuboids, cones, cylinders, long poles, etc.), textures, and sizes. As in \cite{xu2025navrl}, the high diversity of the scene improves the expert’s generalization ability, as well as the diversity of the collected data. Robots and their corresponding goals are initialized at random positions on the terrain, with initial distances of up to 30~m.

\noindent\textbf{Squeezing environments.}  
The static narrow-squeezing scene consists of densely placed pillars randomly distributed across the terrain and walls with narrow, randomized gaps. The expert must navigate safely and smoothly through these passages using visual feedback from four surround-view cameras. For example, observations from the left and right cameras help determine whether the robot can pass between adjacent obstacles. 

\noindent\textbf{Avoiding environments.}  
The dynamic avoidance scene contains densely placed dynamic obstacles moving at velocities between 0.5~m/s and 1.5~m/s, requiring the RL expert to actively avoid collisions. These dynamic obstacles exhibit diverse sizes and geometries, such as spheres, cubes, rods, and cones. Similarly to the reaching scene, the distances between the robot and the target are sampled with values up to 10~m, and no collision occurs when the robot is initialized. 

\noindent\textbf{Simulation setup.}
We construct our environments based on IsaacLab~\cite{mittal2023orbit} for its modular design and strong reinforcement learning support.
Following~\cite{cai2025navdp}, in all three scenes, the robot is abstracted as a cuboid of size \([0.70~\text{m}, 0.35~\text{m}, 0.50~\text{m}]\) in simulation to improve computational and rendering efficiency.


\noindent\textbf{RL experts architecture.} Although trained in three different scenes, all RL experts share a similar training methodology. We employ the Proximal Policy Optimization (PPO)~\cite{schulman2017proximal} algorithm in simulation with 128 parallel robots. The observation at each time step \(t\) is defined as:
\begin{equation}
    O_{t}^{\text{RL}} = [d_t^{\text{front}}, d_t^{\text{right}}, d_t^{\text{back}}, d_t^{\text{left}}, {a}_{t-1}, {g}_t],
\end{equation}
where \(d_t^{\text{front}}, d_t^{\text{right}}, d_t^{\text{back}}, d_t^{\text{left}}\) are the depth images from four cameras mounted on the respective sides of the robot, \({a}_{t-1}\) is the last applied action and \({g}_t\) is the relative position of the goal. As demonstrated in \Cref{fig:pipeline}. Each depth image from the four views is encoded into a feature vector by ResNet-18~\cite{he2016deep}, which is concatenated with the last action \({a}_{t-1}\), the goal position \({g}_t\), and a history token from the last hidden layer of a three-layer MLP. The concatenated features are fed to the MLP to predict the velocity action \({a}_t = [v^\text{x}_t, v^\text{y}_t, v^\text{yaw}_t]\).
The last hidden state of the MLP is stored and passed to the next time step as the history token. The output \({a}_t\) is then multiplied and clipped within the maximum velocity limits \({v}_\text{max} = [1.5~\text{m/s},1.0~\text{m/s},\pi/4.0~\text{rad/s}]\):
\begin{equation}
    {v}_t = \max \{ \min \{{a}_t, 1.0\}, -1.0 \} \times {v}_\text{max}.
\end{equation}
The resulting velocity \({v}_t\) is then applied in the simulation. During training, Gaussian noise is added to proprioceptive velocity observations to improve robustness.

\noindent\textbf{RL training rewards and termination conditions.}
The reward encourages reasonable, goal-directed, and collision-free behavior:
\begin{equation}
    r_\text{Cap.} = \alpha_\text{Cap.}r_{\text{goal}} + \beta_\text{Cap.}r_{\text{step}} + \gamma_\text{Cap.}r_{\text{reg}} + \delta_\text{Cap.}r_{\text{col}},
\label{eq3}
\end{equation}
where $\text{Cap.}$ denotes the reaching, squeezing, and avoiding experts; $r_{\text{goal}}$ rewards progress toward the goal; $r_{\text{step}}$ penalizes each step to encourage efficiency; $r_{\text{reg}}$ regularizes actions to discourage undesirable behaviors (e.g., retreating, circling); and $r_{\text{col}}$ penalizes collisions with obstacles or other robots. To guide and specialize the three RL experts, their reward coefficients $\alpha_\text{Cap.}, \beta_\text{Cap.}, \gamma_\text{Cap.}, \delta_\text{Cap.}$ are designed to differ. Specifically, we set $\beta_\text{Cap.} = -0.05$ and $\delta_\text{Cap.} = -15$ for all experts. For the regularization terms, we assign $\gamma_\text{reaching}=0.05,\ \gamma_\text{squeezing}=0.02,\ \gamma_\text{avoiding}=0$, thereby enforcing stronger regulation in the relatively easier reaching scene while applying weaker or no regularization in the more challenging squeezing and avoiding scenes. Finally, for the goal-reaching reward, we set $\alpha_\text{reaching} = \alpha_\text{avoiding} = 1.2$ and $\alpha_\text{squeezing} = 1.5$, ensuring that the squeezing expert remains decisive when traversing narrow gaps. More details are provided in the supplementary material.

\subsection{Student VLA Model}

\noindent\textbf{Visual Observation Encoding.}  
To support a 360$^\circ$ observation, we extend a video-based VLA model~\cite{zhang2024navid} to support multi-view observations. Given a sliding window (length $k = 8$) of navigation history of four-view RGB images $O_T=I_{T-k+1:T}^N$, which include a front-view, right-view, back-view, and left-view of robots. We then encode all frames into a sequence of visual tokens $E^\text{visual} \in \mathbb{R}^{P \times C}$ ($P=576$ is the patch size and $C$ is the token dimensions) using a visual foundation model (implemented with SigLIP~\cite{zhai2023siglip}) and a cross-modal projector (a two-layer MLP~\cite{liu2023llava}). Similar to previous methods~\cite{zhang2024navid,zhang2024uni}, we conduct grid-based average pooling of the visual tokens, and use fine-grained visual tokens $E^\text{fine} \in \mathbb{R}^{16 \times C}$ for latest observation and coarse-grained visual tokens $E^\text{coarse} \in \mathbb{R}^{4 \times C}$ for navigation history. Finally, we can organize the visual token as:
\begin{equation}
    E_\text{visual} = \{E_{T-k+1}^\text{F/R/B/L\_coarse}, ... , E_{T-1}^\text{F/R/B/L\_coarse}, E_{T}^\text{F/R/B/L\_fine}\},
\end{equation}
where all four camera views (Front, Right, Back, Left) are used at each timestep. The use of a sliding window for token selection helps maintain a reasonable visual token sequence length (192 tokens), thereby ensuring consistent inference speed.

\noindent\textbf{Action Forwarding.}  
With organized visual tokens, we format the relative point goal $g_t$ into a textual prompt and encode the prompt as language tokens $E_\text{text}$. Here, using a textual prompt to represent the point goal allows us to co-train the navigation data with open-world VQA data, which has been widely proven to mitigate the sim-to-real gap~\cite{zhang2024uni,wang2025trackvla}. We then feed the concatenation of the visual tokens $E_\text{visual}$ and language tokens $E_\text{text}$ into a large language model (implemented using Qwen2~\cite{qwen2}) to obtain the predicted action token $E_\text{action}$. Finally, the action token is processed by an action head (implemented with a two-layer MLP) to predict the velocity ($v_t=[v^{\text{x}}_t, v^{\text{y}}_t, v^{\text{yaw}}_t]$) of the robot.

Here, we apply mean-square-error loss $L_\text{action} = \text{MSE}(v^\text{Pred.}, v^\text{GT})$ for action prediction, and retain the cross-entropy loss $L_\text{QA}$ for open-world question-answering data, similar to \cite{wang2025trackvla}. The overall loss is defined as: $L = \beta L_\text{action} + L_\text{QA}$, where the $\beta$ is set to $5$ to balance the magnitude of the two loss terms. Notably, through a tiered architecture that incorporates visual tokens, our method achieves an inference speed of 7 Hz using a 7B-parameter LLM, while maintaining strong performance in challenging scenarios.
This framework could be further improved by integrating acceleration techniques such as quantization~\cite{lang2024comprehensive,frantar2022gptq,lin2024awq}.



\subsection{RL Experts–VLA iteration}

To distill expertise from multiple RL experts into the student VLA model, we design a two-stage training process.

\noindent\textbf{Initial expert data collection and VLA finetuning.} 
To endow the VLA model with initial navigation capability, we first collect a diverse set of trajectories generated by RL experts in simulation. For each expert, we run 64 parallel robots in their respective simulation environments, recording the observations, goals, and corresponding expert actions. Only trajectories that successfully reach the goal are retained, ensuring that erroneous actions from imperfect RL experts are excluded. All retained trajectories are aggregated into a dataset containing 500k steps, each represented as \([O_i, {g}_i, {a}_i]\), where \({a}_i\) is the ground truth velocity. During collection, we randomized the height of the camera within $[0.4\,\text{m},0.5\,\text{m}]$ and the field of view (FOV) of each camera within $[100^\circ,140^\circ]$. The initial dataset also contains 100k real-world VQA data. As shown in \Cref{fig:iteration}, after being trained on the large-scale dataset, the initial VLA model performs well in the reaching scene but poorly in others, and in all cases underperforms the corresponding RL expert.

\begin{figure}[t]
\centering
\includegraphics[width=\linewidth]{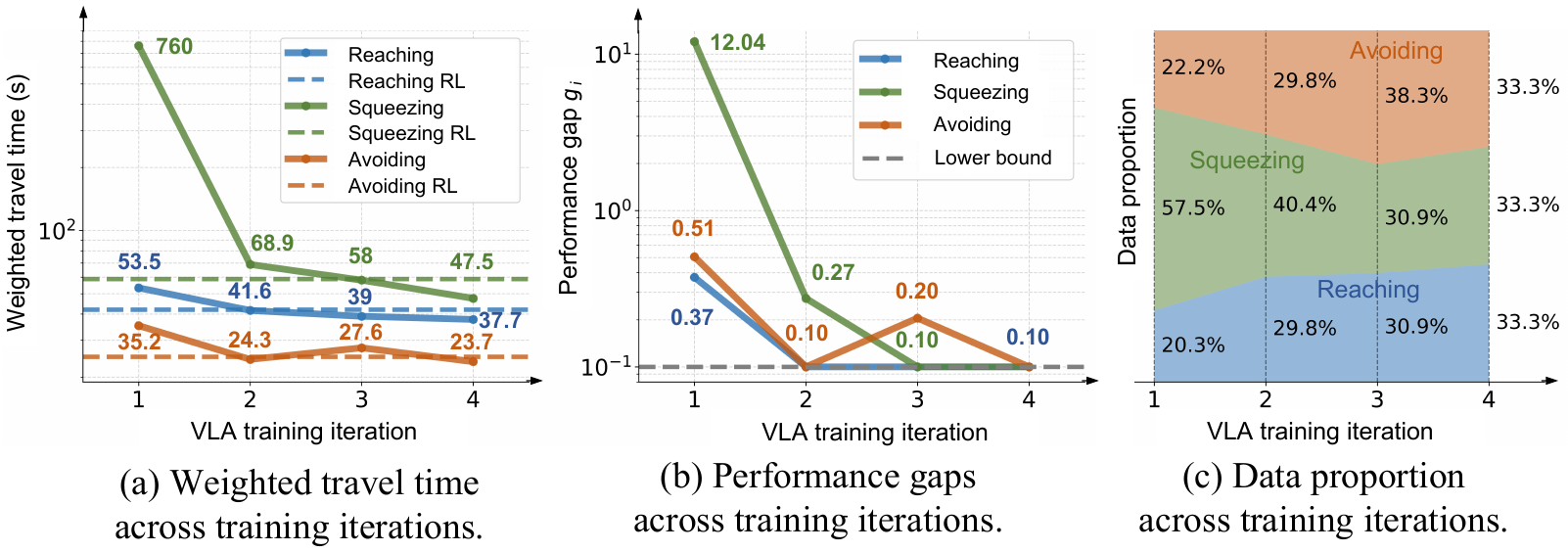}
\caption{
\textbf{Online training dynamics across iterations.}
(a) Weighted travel time (WTT) of the VLA model compared with RL experts.
(b) Performance gap $g_i$ between the VLA model and RL experts.
(c) Data proportion of the online collected data for reaching, squeezing, and avoiding.
With four iterations, shrinking performance gaps lead the capability-balancing module to equalize sampling ratios, and VLA eventually matches or surpasses all experts in WTT, yielding a uniform data ratio.}
\vspace{-1.5mm}
\label{fig:proportion}
\end{figure}

\noindent\textbf{Teachers-student online training iteration.}
We collect online expert data iteratively in a data aggregation manner.
The preliminarily trained VLA model is deployed in the three simulation scenes to collect actions from the corresponding RL teachers. However, the varying complexities of the three capabilities result in uneven performance of VLA across the three scenes, like performing well in the reaching task but underperforming in the more challenging squeezing and avoiding tasks. This motivates the need for a more balanced use of training data from different scenes. To address this issue, we propose a capability-balanced data aggregation method. It leverages weighted travel time (WTT) $W$ (the average time of successful episodes divided by success rate) to measure the performance gap between the VLA model and the RL experts. The gap is defined as: 
\begin{equation}
    g_\text{Cap.} = \max\left\{0, \frac{W_\text{Cap.}^{\text{VLA}} - W_\text{Cap.}^\text{RL}}{W_\text{Cap.}^\text{RL}}\right\} + \epsilon,
\end{equation}
where $W_\text{Cap.}^\text{RL}$ represents the WTT of the reaching, squeezing, and avoiding experts, respectively, and $W_\text{Cap.}^\text{VLA}$ represents the WTT of the current VLA model in each scene. We set $\epsilon$ to $0.1$ to maintain a set of data when the VLA model already outperforms the RL expert. Based on this gap, we compute the data proportion $p_\text{Cap.}$ for each capability as:
\begin{equation}
p_\text{Cap.} = \frac{g_\text{Cap.}^\alpha}{\sum_\text{Cap.} g_\text{Cap.}^\alpha},
\end{equation}
with $\alpha=0.3$ to smooth the distribution. Intuitively, larger performance gaps lead to higher proportions of the corresponding expert data being aggregated, as illustrated in \Cref{fig:proportion}. Finally, using the computed proportions $p_\text{Cap.}$, we aggregate the expert datasets in a capability-balanced manner and fine-tune the VLA. This enables the transfer of additional collision-avoidance and navigation skills. After fine-tuning, the VLA is re-evaluated and the process is repeated iteratively with updated $p_\text{Cap.}$ values until no further improvement is observed.

\subsection{Implementation Details}

\noindent\textbf{RL Training strategy.}
Each RL expert is trained in IsaacLab~\cite{mittal2023orbit} on an NVIDIA RTX 4090 GPU for 8-12 hours, using $N=128$ parallel environments. 
The policy adopts a history-aware actor-critic architecture. Both the actor and critic are implemented as three-layer MLPs with hidden dimensions of [512, 256, 128] and use the ELU activation function~\cite{clevert2015fast}. Regarding the sensory input, the depth values from the cameras are clipped to [0.01m, 4.0m] to filter out noise. 
The action distribution is initialized with a noise standard deviation of 0.2.

\noindent\textbf{VLA training strategy.} Our initial VLA model is fine-tuned on 8 NVIDIA H100 GPUs for about 5 hours, totaling 40 GPU hours. We leverage the pre-trained weights of both visual encoders (SigLIP~\cite{zhai2023siglip}) and LLM (Qwen2-7B~\cite{qwen2}). The initial training contains 500k steps from three RL experts and 100k visual question answering (VQA) data. For VQA data, we use a subset of the dataset in ~\cite{shen2024longvu} and the frames are sampled at 1 FPS, following ~\cite{zhang2024navid}. Each iteration of teacher-student training costs about 2 hours and contains 200k steps of online collected expert data and 40k VQA data. 

\noindent\textbf{Deployment strategy.} 
We deploy our method on the Unitree GO2 robot. The VLA model is executed on a server equipped with an NVIDIA RTX 5090 GPU, while the onboard computer on the robot continuously sends requests and receives velocity commands. We use four fisheye cameras mounted on the front, right, back, and left sides of the robot to obtain the four-view real-time images, which are then undistorted into normal perspective images and fed into the VLA model. Given the velocity $v_t$ inferred by the VLA model, the low-level controller follows the velocity until the next response. The average response fps is about 7 Hz, which is sufficient for our navigation task.

\section{Experiments}

\subsection{Experiment Setup}
\noindent\textbf{Simulation environment setup.}
For comparative evaluation, we first assess MM-Nav on the public InternVLA-N1 System-1 (S1) point-goal navigation benchmark~\cite{wang2025internvla}. Following the benchmark protocol, we evaluate on 40 scenes from InternScenes~\cite{zhong2025internscenes}, using the same 100 start-goal pairs as InternVLA-N1 and the Dingo robot platform with omnidirectional control. To further evaluate our method in more challenging settings, we additionally construct four custom test environments in IsaacLab~\cite{mittal2023orbit}. Three fixed capability-specific scenes are designed to separately assess the three target capabilities, namely reaching, squeezing, and avoiding. In addition, we build a \texttt{Mixed} scene that combines static obstacles, dynamic obstacles, and narrow passages, all of which lie outside the VLA model's training distribution. All objects in the \texttt{Mixed} scene are rendered with materials visually distinct from those used during training, and the difficulty of the squeezing segment is intentionally reduced to keep the overall evaluation balanced. Each episode terminates when the robot reaches the goal, collides with an obstacle, or exceeds the time limit. The timeout is set to 90 seconds for the three capability-specific scenes and extended to 120 seconds for the more complex \texttt{Mixed} scene. Each method was evaluated over 100 episodes per scene following its original configuration.
\begin{table}[t]
\caption{Results in InternVLA-N1~\cite{wang2025internvla} System-1 point-goal navigation benchmark. For fair comparison with prior front-view baselines, we report a single-view variant.}
\label{tab:sim_results}
\centering
\scalebox{0.85}{
\setlength{\tabcolsep}{1mm}
\renewcommand{\arraystretch}{1.15}
\begin{tabular}{l|c|cc|cc}
\toprule
\multirow{2}{*}{Methods} & \multirow{2}{*}{Obs.} & \multicolumn{2}{c|}{Home} & \multicolumn{2}{c}{Commercial} \\
                         &                        & SR$\uparrow$ & SPL$\uparrow$ & SR$\uparrow$ & SPL$\uparrow$ \\
\midrule
DD-PPO~\cite{wijmans2020ddppo}         & RGB-D & 0.4  & 0.4  & 5.3  & 5.2  \\
iPlanner~\cite{yang2023iplanner}       & Depth & 43.0 & 40.6 & 54.6 & 52.8 \\
ViPlanner~\cite{roth2024viplanner}     & RGB-D & 45.0 & 43.2 & 63.7 & 61.9 \\
LoGoPlanner~\cite{peng2025logoplanner} & RGB-D & 57.3 & 52.4 & 67.1 & 63.9 \\
InternVLA-N1(S1)~\cite{wang2025internvla} & RGB-D & 60.0 & 55.6 & 71.4 & 68.2 \\
NavDP~\cite{cai2025navdp}              & RGB-D & 60.3 & 54.7 & 74.1 & 70.5 \\
SIDP~\cite{zhang2026self}              & RGB-D & 63.2 & 56.5 & 81.2 & 73.4 \\
Mixed-RL &Depth & 22.0 & 10.4 & 26.9 & 22.8\\
\rowcolor{mygray}
Ours (without VQA)  & RGB   & 76.4 & 73.7 & 73.9 & 72.6 \\\rowcolor{mygray}
Ours (single view)  & RGB   & 79.9 & 77.6 & 86.7 & 84.7 \\
\rowcolor{mygray}
\textbf{Ours (multi-view)} & RGB & \textbf{86.3} & \textbf{81.1} & \textbf{89.9} & \textbf{85.5} \\
\bottomrule
\end{tabular}
}
\end{table}

\begin{figure}[t]
    \centering
    \includegraphics[width=\linewidth]{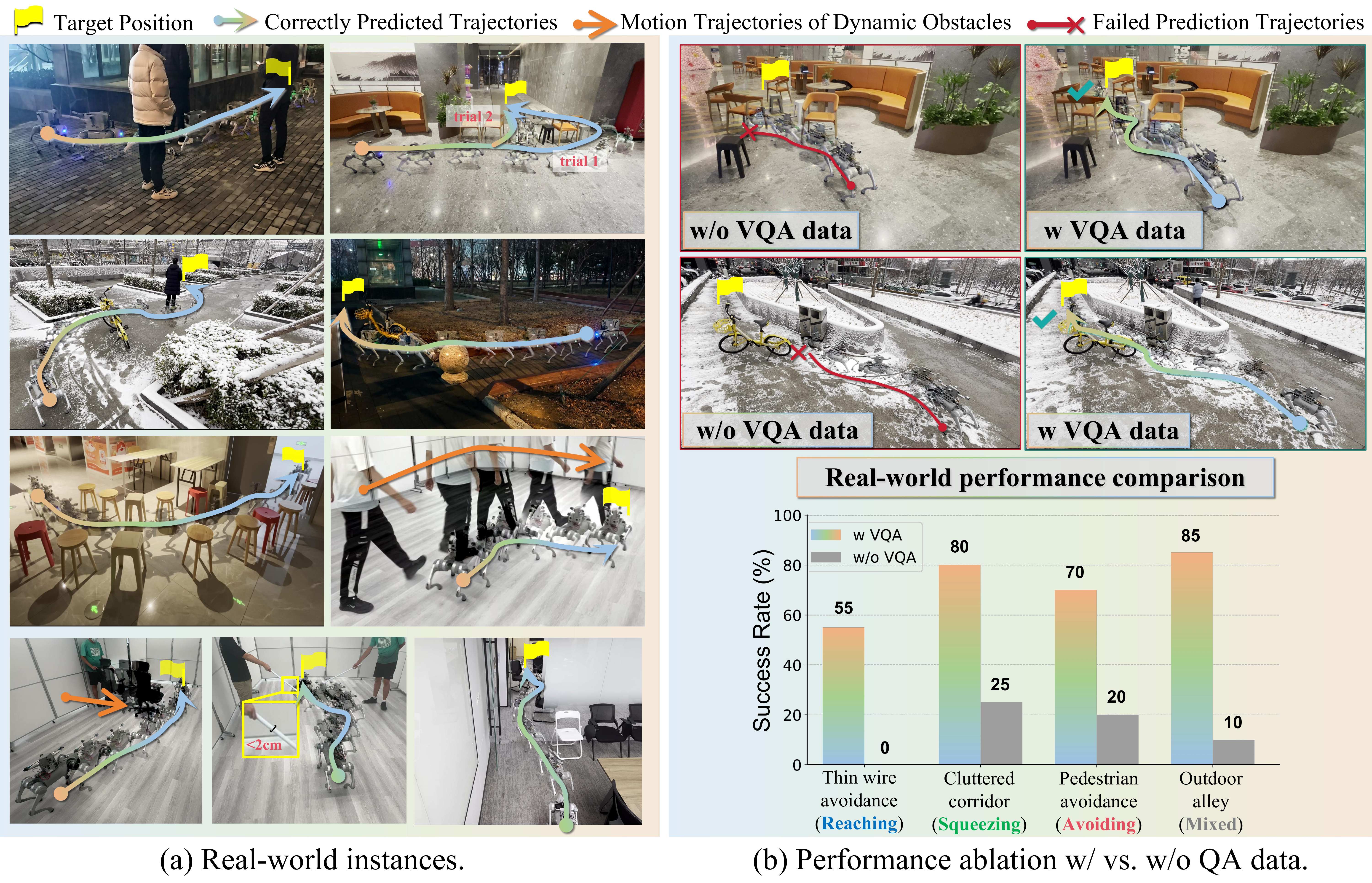}
    
    \caption{
        \textbf{Real-world Experiments.}
        (a): Representative deployments of \ours in diverse real-world environments under varying scene layouts and lighting conditions.
        (b): A real-world ablation study on VQA co-training.
    }

    \label{fig:galary}
    \vspace{-2mm}

\end{figure}

\noindent\textbf{Real-world environment setup.}
\label{realworldenv}
We conduct extensive real-world deployments across diverse indoor and outdoor environments to evaluate its sim-to-real transfer and generalization ability under varied layouts, objects, and lights. From these trials, we select four representative scenarios for detailed analysis. The first two scenarios are shown in \Cref{fig:teaser}: \textit{Thin Wire Avoidance}, a corridor scene densely populated with many suspended thin wires, and \textit{Cluttered Corridor}, a narrow office corridor filled with obstacles and partially transparent glass walls. The other two scenarios are shown in \Cref{fig:galary}: \textit{Pedestrian Avoidance}, which involves pedestrians moving across and near the robot's forward path, and \textit{Outdoor Alley}, a nighttime outdoor scene containing previously unseen obstacles such as a stone bollard and a bicycle. These examples are chosen to show how the capabilities learned in simulation transfer to real-world navigation.

\noindent\textbf{Metrics and baselines.}
On the public InternVLA-N1 System-1 benchmark, we report two standard metrics: Success Rate ({SR}) and Success weighted by Path Length ({SPL}). On our custom IsaacLab benchmark, we report three metrics: (1) Success Rate ({SR}), (2) Collision Rate ({CR}), and (3) Weighted Travel Time ({WTT}), where {WTT} is defined as the average completion time of successful episodes divided by the SR. For the IsaacLab evaluation, each method is rolled out for 100 episodes in each scene. We compare MM-Nav against DD-PPO, iPlanner, ViPlanner, LoGoPlanner, InternVLA-N1(S1), NavDP, and SIDP on the public benchmark, and against iPlanner, ViPlanner, and NavDP in the custom IsaacLab environments.

\subsection{Quantitative Results}

The results of \ours compared to other baselines on the InternVLA-N1 S1 benchmark are shown in \Cref{tab:sim_results}. Our method achieves the highest performance across both Home and Commercial splits, significantly outperforming prior RGB-D and depth-based approaches, despite relying solely on RGB observations. In particular, \ours improves the SR by approximately 25\% over the strongest baseline. However, our RL expert itself performs poorly on this benchmark, suffering from severe domain gap in different simulation environments. These results demonstrate that our multi-capability training paradigm effectively enhances generalization across diverse environments.

\begin{table}[t]\centering
\caption{
Quantitative comparison on IsaacLab simulator in three capability-specific scenes and a mixed scene. Each method was evaluated over 100 episodes per scene.}
\scalebox{0.85}{ 
\setlength{\tabcolsep}{1mm} 
\begin{tabular}{lcccccccccccc}
\toprule
\multirow{2}{*}[-0.5ex]{\textbf{Methods}} & \multicolumn{3}{c}{\textcolor{reaching}{Reaching}} 
        & \multicolumn{3}{c}{\textcolor{squeezing}{Squeezing}} 
        & \multicolumn{3}{c}{\textcolor{avoiding}{Avoiding}} 
        & \multicolumn{3}{c}{\textcolor{reaching}{M}\textcolor{squeezing}{ix}\textcolor{avoiding}{ed}} \\
\cmidrule(lr){2-4}\cmidrule(lr){5-7}\cmidrule(lr){8-10}\cmidrule(lr){11-13}
{} & {SR$\uparrow$} & {CR$\downarrow$} & {WTT$\downarrow$} 
   & {SR$\uparrow$} & {CR$\downarrow$} & {WTT$\downarrow$} 
   & {SR$\uparrow$} & {CR$\downarrow$} & {WTT$\downarrow$} 
   & {SR$\uparrow$} & {CR$\downarrow$} & {WTT$\downarrow$} \\
\midrule
{iPlanner~\cite{yang2023iplanner}}   & 19 & 81 & 93.4  & 2 & 94 & 881.25  & 15 & 85 & 73.4 & 4 & 94 & 736.9  \\
{ViPlanner~\cite{roth2024viplanner}} & 43 & 57 & 42.2  & 4 & 96 & 452.8  & 22 & 78 & 36.36  & 18 & 82 & 215.4  \\
{NavDP~\cite{cai2025navdp}}          & 69 & 31 & \textbf{27.3}  & 18 & 82 & 115.4  & 27 & 73 & 30.0  & 23 & 77 & 178.6  \\
\rowcolor{mygray}
\textbf{Ours}     & \textbf{80} & \textbf{20} & 31.0  & \textbf{71} & \textbf{19} & \textbf{42.2 } & \textbf{68} & \textbf{32} & \textbf{20.9 } & \textbf{47} & \textbf{26} & \textbf{127.5 } \\
\bottomrule
\end{tabular}
}
\vspace{-1mm}
\label{tab:benchmark}
\end{table}

The simulation benchmark results on our custom IsaacLab benchmark are summarized in \Cref{tab:benchmark}. Overall, \ours achieves the highest SR, the lowest CR, and competitive or shortest WTT across almost all scenes, indicating both safe and efficient navigation. Among the baselines, NavDP~\cite{cai2025navdp} performs competitively in the Reaching scene but is fundamentally constrained by its single front-facing camera, which limits its field of view (FOV). This restriction leads to characteristic failures in complex scenarios, particularly when obstacles may approach from lateral or rear directions or remain partially occluded. ViPlanner~\cite{roth2024viplanner} and iPlanner~\cite{yang2023iplanner} suffer from similar FOV limitations and exhibit less responsive control behaviors, resulting in lower performance. In contrast, our 360$^\circ$ surround-view perception combined with continuous velocity control enables more robust behavior in cluttered, dynamic and constrained environments. The visualization results in simulators are provided in the supplementary material.

\subsection{Qualitative Results}
Across the four selected real-world scenarios, \ours demonstrates strong zero-shot sim-to-real transfer. In \textit{Thin Wire Avoidance}, the robot successfully avoids these thin obstacles, which are challenging for LiDAR-based systems to reliably perceive~\cite{wang2025omni}, indicating that the learned visual representation captures subtle obstacle cues and supports the \textit{reaching} capability learned in simulation. In \textit{Cluttered Corridor}, the robot navigates through the cluttered passage while handling unseen obstacles and challenging materials like glass, reflecting the \textit{squeezing} capability learned in simulation. In \textit{Pedestrian Avoidance}, the robot adjusts its actions online according to the pedestrians' trajectories and precisely avoids close-range interactions while maintaining progress toward the goal, demonstrating the learned \textit{avoiding} capability. In \textit{Outdoor Alley}, the robot accurately avoids these previously unseen obstacles under low-light conditions, further demonstrating robust real-world generalization.

\subsection{Ablation Study}


\noindent\textbf{Generalization ability gained from VQA data.}
We conduct an ablation study on co-training VQA data to prove that large-scale real world VQA data is essential in helping model generalize to different simulation environments and bridge sim-to-real gap. The comparison result in unseen simulation environments is shown in \Cref{tab:sim_results}, the VLA model trained without VQA data suffers a large performance drop of more than 20\% in both Home and Commercial scenes. Moreover, we evaluate the SR of \ours with and without VQA data across several challenging real-world scenarios mentioned in \Cref{realworldenv}. Each model is evaluated for 20 trials in every scenario. As shown in \Cref{fig:galary}~(b), the lack of VQA data leads to a very poor performance, unable to correctly avoid obstacles with thin or irregular shape. While the VLA model co-trained with VQA data demonstrates strong sim-to-real transfer ability.
\begin{wrapfigure}{r}{0.64\textwidth}
\vspace{-1.0em}
\centering
\includegraphics[width=\linewidth]{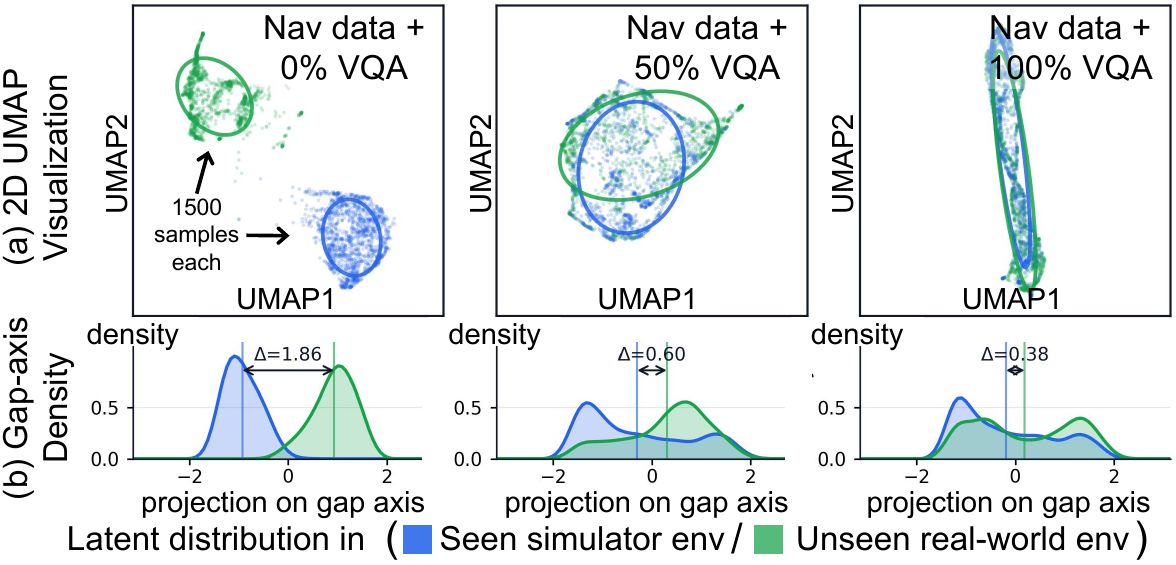}
\vspace{-1.6em}
\caption{Latent distributions under different VQA ratios.}
\label{fig:rebuttal}
\vspace{-1.0em}
\end{wrapfigure}
Beyond the above performance ablations, we further analyze how VQA co-training affects navigation action prediction by visualizing the final-layer VLM hidden states used by the action head. Specifically, we feed synthetic and real-world observations into MM-Nav variants trained with 0\%, 50\%, and 100\% VQA data, and analyze their action-relevant VLM hidden states using UMAP and gap-axis projection in \Cref{fig:rebuttal}. Without VQA co-training, the simulation and real-world latents form clearly separated clusters, indicating a large sim-to-real visual gap. As the amount of VQA data increases, the two domains become progressively better aligned, with the gap-axis distance decreasing from 1.86 to 0.38. These results suggest that VQA co-training exposes the shared VLM backbone to real-image statistics, including lighting variations and complex scene structures, thereby helping align real observations with the simulation-trained latent space. 

\noindent\textbf{Multi-view vs. single-view observations.}
We further study the impact of our 360$^\circ$ surround-view input by training a single-view variant that uses only the front-facing RGB stream, while keeping the model, training data, and optimization settings unchanged. On the InternVLA-N1 S1 benchmark, multi-view consistently improves performance over single-view, yielding an average gain of +4.8 SR and +2.2 SPL across the Home and Commercial splits (\Cref{tab:sim_results}). This validates that the additional side and rear views provide stronger spatial awareness beyond front-view-only policies, mitigating failures caused by limited FOV under clutter, occlusions, and tight passages.

\begin{figure}[t]
\centering
\includegraphics[width=\linewidth]
{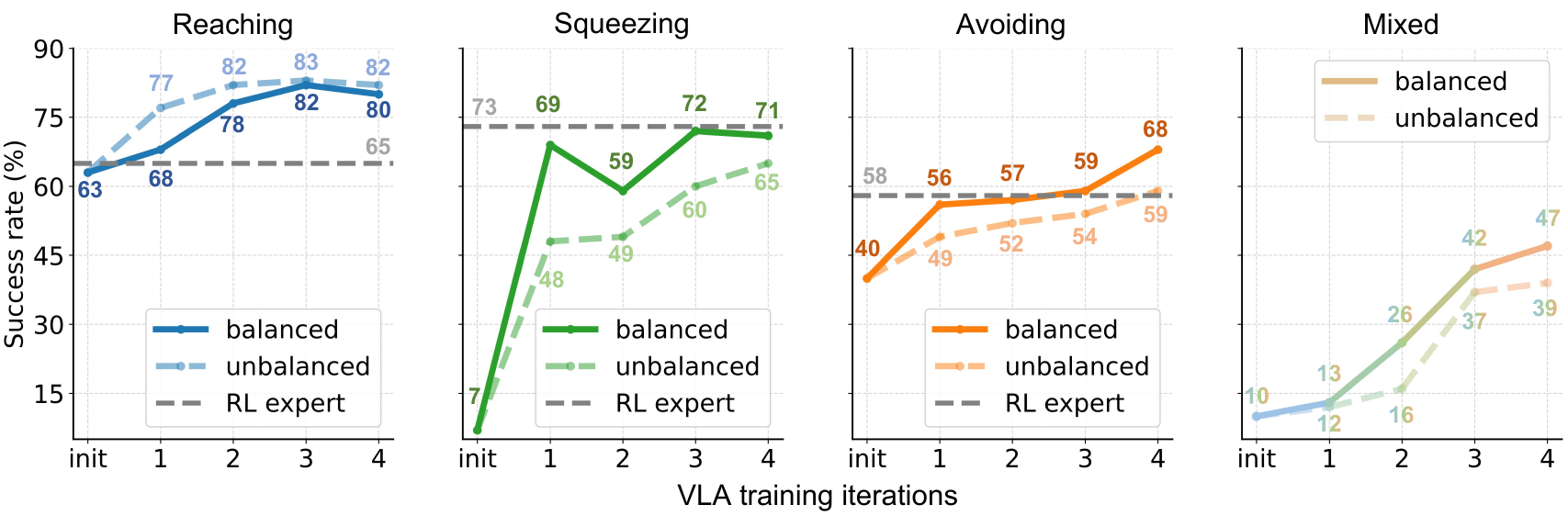}
\caption{Success rate of the VLA model after each iteration.}
\label{fig:iteration}
\vspace{-3mm}
\end{figure}

\noindent\textbf{Performance gains across online training iterations.} The performance of the initial VLA model and its variants after the first four training iterations is evaluated in simulation, as shown in \Cref{fig:iteration}. After the initial behavior-cloning training, the VLA model shows clear performance gaps across all three capabilities, particularly in squeezing. In the first iteration, the capability-balanced data aggregation method places more emphasis on squeezing, leading to substantial gains. As a result, the VLA model after the first iteration becomes comparable to, or even surpasses, the RL experts. In the second and third iterations, the reaching capability is further improved, although the VLA model had already exceeded the corresponding RL expert. This may be attributed to better integration of the squeezing and avoiding capabilities, which also benefits the relatively simpler reaching task. After the fourth iteration, performance across the three tasks converges, and no further iterations are conducted.

\noindent\textbf{Capability-balanced data aggregation method.}
Comparison experiments are conducted to evaluate the effectiveness of our capability-balanced data aggregation method. Starting from the same initial VLA model, we train one set of iterations with balanced data and another with unbalanced data (\Cref{fig:iteration}), while keeping the total number of samples per iteration the same. The results show that the capability-balanced method can complement underdeveloped capabilities in time, leading to faster and more stable training.
Although the unbalanced setting achieves better performance on the reaching task, it fails to efficiently improve squeezing and avoiding. Overall, the capability-balanced method better integrates data from different RL experts and prevents the VLA model from overlooking specific capabilities.

\noindent\textbf{Experts Combination strategy.}
We investigate how combining three capability-specific RL teachers improves the VLA student. We train three VLA variants with the same total data budget, each using trajectories from one expert, and also train a single RL expert in the mixed scene requiring all three capabilities. 
\begin{wraptable}{r}{0.5\linewidth}
\vspace{-8pt}
\centering
\caption{\textbf{Comparison of Individual Navigation Capability.} We report the SR (100 episodes) of VLA models and RL experts trained on either single or mixed datasets. The best overall result is highlighted in \textbf{bold}, and the best result within each category is indicated with \underline{underline}.}
\setlength{\tabcolsep}{2mm}
\renewcommand{\arraystretch}{1.0}
\resizebox{0.5\textwidth}{!}{
\begin{tabular}{lccc}
\toprule
\rowcolor{white}
{Methods} & {\textcolor{reaching}{Reaching}} & {\textcolor{squeezing}{Squeezing}} & {\textcolor{avoiding}{Avoiding}} \\
\midrule
{\textcolor{black}{Reaching-RL}}   & \underline{65} & 2 & 33 \\
{\textcolor{black}{Squeezing-RL}}  & 30 & \underline{73} & 19 \\
{\textcolor{black}{Avoiding-RL}}   & 59 & 0 & \underline{58} \\
{\textcolor{black}{Mix}\textcolor{black}{ed}\textcolor{black}{-RL}} & 45 & 58 & 42 \\
\midrule
{\textcolor{black}{Reaching-VLA}}  & 70 & 4 & 35 \\
{\textcolor{black}{Squeezing-VLA}} & 44 & 64 & 28 \\
{\textcolor{black}{Avoiding-VLA}}  & 63 & 0 & 62 \\
{\textcolor{black}{Mix}\textcolor{black}{ed}\textcolor{black}{-VLA}} & \underline{\textbf{80}} & \underline{\textbf{71}} & \underline{\textbf{68}} \\
\bottomrule
\end{tabular}
}
\label{tab:cross_scene}
\vspace{-10pt}
\end{wraptable} As shown in \Cref{tab:cross_scene}, the mixed-scene RL expert covers all capabilities but cannot match any specialized expert, leading to compromised performance. It also performs poorly on the InternVLA-N1 System-1 point-goal navigation benchmark (\Cref{tab:benchmark}), further indicating the weak generalization ability of RL policies beyond their training distribution. Likewise, single-expert VLA variants perform well in-domain but generalize poorly to unseen capabilities. For example, training only on squeezing data may bias the model toward static obstacle interactions, while weakening its ability to actively avoid dynamic obstacles. In contrast, the mixed-data VLA model achieves stronger cross-capability performance, suggesting that different capabilities are complementary and help the student learn more transferable representations.

\section{Conclusion}
We have presented \ours, a multi-view VLA model that acquires robust visual navigation skills from a collective of specialized RL experts. Our training process consists of two key stages: first, an initial VLA finetuning phase where a student policy learns from a large offline dataset collected from the RL teachers; second, an online teachers-student training iteration where the student is deployed in simulation to receive on-the-fly, capability-balanced supervision for further refinement. \ours demonstrates strong sim-to-real transfer and ultimately outperforms its RL experts, proving the synergistic benefit of learning multiple capabilities. \ours provides a scalable and effective blueprint for training a new generation of general-purpose visual navigation agents. Future work includes investigating the cross-embodiment potential of our training strategy and further advancing visual-only navigation.

\section*{Acknowledgements}

This work was supported by institutional research resources and computing platforms from Peking University and Galbot.
The authors are also grateful to the members of PKU-EPIC and Galbot for helpful discussions, infrastructure support, and assistance with real-world robot deployment. 

\bibliographystyle{splncs04}
\bibliography{main}

@String(ICCV  = {Int. Conf. Comput. Vis.})

@String(NeurIPS = {Adv. Neural Inform. Process. Syst.})

@String(ICLR  = {Int. Conf. Learn. Represent.})

@String(JMLR  = {J. Mach. Learn. Res.})

@String(ICCV  = {ICCV})

@String(NeurIPS = {NeurIPS})

@String(ICLR  = {ICLR})

@String(JMLR  = {JMLR})

@inproceedings{shah2023gnm,
  title={Gnm: A general navigation model to drive any robot},
  author={Shah, Dhruv and Sridhar, Ajay and Bhorkar, Arjun and Hirose, Noriaki and Levine, Sergey},
  booktitle={2023 IEEE International Conference on Robotics and Automation (ICRA)},
  pages={7226--7233},
  year={2023},
  organization={IEEE}
}

@inproceedings{shah2023lm,
  title={Lm-nav: Robotic navigation with large pre-trained models of language, vision, and action},
  author={Shah, Dhruv and Osi{\'n}ski, B{\l}a{\.z}ej and Levine, Sergey and others},
  booktitle={Conference on robot learning},
  pages={492--504},
  year={2023},
  organization={PMLR}
}

@inproceedings{zhou2025navgpt,
  title={Navgpt-2: Unleashing navigational reasoning capability for large vision-language models},
  author={Zhou, Gengze and Hong, Yicong and Wang, Zun and Wang, Xin Eric and Wu, Qi},
  booktitle={European Conference on Computer Vision},
  pages={260--278},
  year={2025},
  organization={Springer}
}

@String(ICCV= {Int. Conf. Comput. Vis.})

@String(NeurIPS= {Adv. Neural Inform. Process. Syst.})

@String(ICLR = {Int. Conf. Learn. Represent.})

@String(NeurIPS  = {NeurIPS})

@inproceedings{lang2024comprehensive,
  title={A comprehensive study on quantization techniques for large language models},
  author={Lang, Jiedong and Guo, Zhehao and Huang, Shuyu},
  booktitle={2024 4th International Conference on Artificial Intelligence, Robotics, and Communication (ICAIRC)},
  pages={224--231},
  year={2024},
  organization={IEEE}
}

@inproceedings{he2016deep,
  title={Deep residual learning for image recognition},
  author={He, Kaiming and Zhang, Xiangyu and Ren, Shaoqing and Sun, Jian},
  booktitle={Proceedings of the IEEE conference on computer vision and pattern recognition},
  pages={770--778},
  year={2016}
}

@article{eftekhar2024one,
  title={The One RING: a Robotic Indoor Navigation Generalist},
  author={Eftekhar, Ainaz and Weihs, Luca and Hendrix, Rose and Caglar, Ege and Salvador, Jordi and Herrasti, Alvaro and Han, Winson and VanderBil, Eli and Kembhavi, Aniruddha and Farhadi, Ali and others},
  journal={arXiv preprint arXiv:2412.14401},
  year={2024}
}

@inproceedings{ross2011reduction,
  title={A reduction of imitation learning and structured prediction to no-regret online learning},
  author={Ross, St{\'e}phane and Gordon, Geoffrey and Bagnell, Drew},
  booktitle={Proceedings of the fourteenth international conference on artificial intelligence and statistics},
  pages={627--635},
  year={2011},
  organization={JMLR Workshop and Conference Proceedings}
}

@article{huang2022visual,
  title={Visual Language Maps for Robot Navigation},
  author={Huang, Chenguang and Mees, Oier and Zeng, Andy and Burgard, Wolfram},
  journal={arXiv preprint arXiv:2210.05714},
  year={2022}
}

@inproceedings{wijmans2020ddppo,
  title = {{DD-PPO}: {L}earning Near-Perfect PointGoal Navigators from 2.5 Billion Frames},
  author =  {Erik Wijmans and Abhishek Kadian and Ari Morcos and Stefan Lee and Irfan Essa and Devi Parikh and Manolis Savva and Dhruv Batra},
  booktitle = {International Conference on Learning Representations (ICLR)},
  year =    {2020}
}

@article{zhang2024navid,
  title={Navid: Video-based vlm plans the next step for vision-and-language navigation},
  author={Zhang, Jiazhao and Wang, Kunyu and Xu, Rongtao and Zhou, Gengze and Hong, Yicong and Fang, Xiaomeng and Wu, Qi and Zhang, Zhizheng and Wang, He},
  journal={Robotics: Science and Systems},
  year={2024}
}

@article{zhang2024uni,
    title={Uni-NaVid: A Video-based Vision-Language-Action Model for Unifying Embodied Navigation Tasks},
    author={Zhang, Jiazhao and Wang, Kunyu and Wang, Shaoan and Li, Minghan and Liu, Haoran and Wei, Songlin and Wang, Zhongyuan and Zhang, Zhizheng and Wang, He},
    journal={Robotics: Science and Systems},
    year={2025}
}

@article{zhou2023navgpt,
  title={NavGPT: Explicit Reasoning in Vision-and-Language Navigation with Large Language Models},
  author={Zhou, Gengze and Hong, Yicong and Wu, Qi},
  journal={arXiv preprint arXiv:2305.16986},
  year={2023}
}

@inproceedings{song2023llm,
  title={Llm-planner: Few-shot grounded planning for embodied agents with large language models},
  author={Song, Chan Hee and Wu, Jiaman and Washington, Clayton and Sadler, Brian M and Chao, Wei-Lun and Su, Yu},
  booktitle={Proceedings of the IEEE/CVF International Conference on Computer Vision},
  pages={2998--3009},
  year={2023}
}

@inproceedings{liu2023llava,
    author      = {Liu, Haotian and Li, Chunyuan and Wu, Qingyang and Lee, Yong Jae},
    title       = {Visual Instruction Tuning},
    booktitle   = {NeurIPS},
    year        = {2023}
}

@article{zhu2023chatgpt,
  title={Chatgpt asks, blip-2 answers: Automatic questioning towards enriched visual descriptions},
  author={Zhu, Deyao and Chen, Jun and Haydarov, Kilichbek and Shen, Xiaoqian and Zhang, Wenxuan and Elhoseiny, Mohamed},
  journal={arXiv preprint arXiv:2303.06594},
  year={2023}
}

@article{zeng2024poliformer,
  title={PoliFormer: Scaling On-Policy RL with Transformers Results in Masterful Navigators},
  author={Zeng, Kuo-Hao and Zhang, Zichen and Ehsani, Kiana and Hendrix, Rose and Salvador, Jordi and Herrasti, Alvaro and Girshick, Ross and Kembhavi, Aniruddha and Weihs, Luca},
  journal={arXiv preprint arXiv:2406.20083},
  year={2024}
}

@article{long2024instructnav,
  title={InstructNav: Zero-shot System for Generic Instruction Navigation in Unexplored Environment},
  author={Long, Yuxing and Cai, Wenzhe and Wang, Hongcheng and Zhan, Guanqi and Dong, Hao},
  journal={arXiv preprint arXiv:2406.04882},
  year={2024}
}

@inproceedings{yokoyama2024vlfm,
  title={Vlfm: Vision-language frontier maps for zero-shot semantic navigation},
  author={Yokoyama, Naoki and Ha, Sehoon and Batra, Dhruv and Wang, Jiuguang and Bucher, Bernadette},
  booktitle={2024 IEEE International Conference on Robotics and Automation (ICRA)},
  pages={42--48},
  year={2024},
  organization={IEEE}
}

@inproceedings{sridhar2024nomad,
  title={Nomad: Goal masked diffusion policies for navigation and exploration},
  author={Sridhar, Ajay and Shah, Dhruv and Glossop, Catherine and Levine, Sergey},
  booktitle={2024 IEEE International Conference on Robotics and Automation (ICRA)},
  pages={63--70},
  year={2024},
  organization={IEEE}
}

@article{kim2024openvla,
  title={Openvla: An open-source vision-language-action model},
  author={Kim, Moo Jin and Pertsch, Karl and Karamcheti, Siddharth and Xiao, Ted and Balakrishna, Ashwin and Nair, Suraj and Rafailov, Rafael and Foster, Ethan and Lam, Grace and Sanketi, Pannag and others},
  journal={arXiv preprint arXiv:2406.09246},
  year={2024}
}

@inproceedings{roth2024viplanner,
  title={Viplanner: Visual semantic imperative learning for local navigation},
  author={Roth, Pascal and Nubert, Julian and Yang, Fan and Mittal, Mayank and Hutter, Marco},
  booktitle={2024 IEEE International Conference on Robotics and Automation (ICRA)},
  pages={5243--5249},
  year={2024},
  organization={IEEE}
}

@inproceedings{zhai2023siglip,
  title={Sigmoid loss for language image pre-training},
  author={Zhai, Xiaohua and Mustafa, Basil and Kolesnikov, Alexander and Beyer, Lucas},
  booktitle={Proceedings of the IEEE/CVF international conference on computer vision},
  pages={11975--11986},
  year={2023}
}

@article{qwen2,
    title   = {Qwen2 Technical Report}, 
    author  = {An Yang and Baosong Yang and Binyuan Hui and Bo Zheng and Bowen Yu and Chang Zhou and Chengpeng Li and Chengyuan Li and Dayiheng Liu and Fei Huang and Guanting Dong and Haoran Wei and Huan Lin and Jialong Tang and Jialin Wang and Jian Yang and Jianhong Tu and Jianwei Zhang and Jianxin Ma and Jin Xu and Jingren Zhou and Jinze Bai and Jinzheng He and Junyang Lin and Kai Dang and Keming Lu and Keqin Chen and Kexin Yang and Mei Li and Mingfeng Xue and Na Ni and Pei Zhang and Peng Wang and Ru Peng and Rui Men and Ruize Gao and Runji Lin and Shijie Wang and Shuai Bai and Sinan Tan and Tianhang Zhu and Tianhao Li and Tianyu Liu and Wenbin Ge and Xiaodong Deng and Xiaohuan Zhou and Xingzhang Ren and Xinyu Zhang and Xipin Wei and Xuancheng Ren and Yang Fan and Yang Yao and Yichang Zhang and Yu Wan and Yunfei Chu and Yuqiong Liu and Zeyu Cui and Zhenru Zhang and Zhihao Fan},
    journal = {arXiv preprint arXiv:2407.10671},
    year    = {2024}
}

@article{shen2024longvu,
  title={Longvu: Spatiotemporal adaptive compression for long video-language understanding},
  author={Shen, Xiaoqian and Xiong, Yunyang and Zhao, Changsheng and Wu, Lemeng and Chen, Jun and Zhu, Chenchen and Liu, Zechun and Xiao, Fanyi and Varadarajan, Balakrishnan and Bordes, Florian and others},
  journal={arXiv preprint arXiv:2410.17434},
  year={2024}
}

@article{cai2025navdp,
  title={NavDP: Learning Sim-to-Real Navigation Diffusion Policy with Privileged Information Guidance},
  author={Cai, Wenzhe and Peng, Jiaqi and Yang, Yuqiang and Zhang, Yujian and Wei, Meng and Wang, Hanqing and Chen, Yilun and Wang, Tai and Pang, Jiangmiao},
  journal={arXiv preprint arXiv:2505.08712},
  year={2025}
}

@article{wang2025trackvla,
  title={Trackvla: Embodied visual tracking in the wild},
  author={Wang, Shaoan and Zhang, Jiazhao and Li, Minghan and Liu, Jiahang and Li, Anqi and Wu, Kui and Zhong, Fangwei and Yu, Junzhi and Zhang, Zhizheng and Wang, He},
  journal={arXiv preprint arXiv:2505.23189},
  year={2025}
}

@misc{chiang2023vicuna,
    title = {Vicuna: An Open-Source Chatbot Impressing GPT-4 with 90\%* ChatGPT Quality},
    url = {https://lmsys.org/blog/2023-03-30-vicuna/},
    author = {Chiang, Wei-Lin and Li, Zhuohan and Lin, Zi and Sheng, Ying and Wu, Zhanghao and Zhang, Hao and Zheng, Lianmin and Zhuang, Siyuan and Zhuang, Yonghao and Gonzalez, Joseph E. and Stoica, Ion and Xing, Eric P.},
    month = {March},
    year = {2023}
}

@article{wang2025x,
  title={X-Nav: Learning End-to-End Cross-Embodiment Navigation for Mobile Robots},
  author={Wang, Haitong and Tan, Aaron Hao and Fung, Angus and Nejat, Goldie},
  journal={arXiv preprint arXiv:2507.14731},
  year={2025}
}

@article{liu2025compass,
  title={COMPASS: Cross-embodiment Mobility Policy via Residual RL and Skill Synthesis},
  author={Liu, Wei and Zhao, Huihua and Li, Chenran and Biswas, Joydeep and Pouya, Soha and Chang, Yan},
  journal={arXiv preprint arXiv:2502.16372},
  year={2025}
}

@article{kim2025enhancing,
  title={Enhancing Safety of Foundation Models for Visual Navigation through Collision Avoidance via Repulsive Estimation},
  author={Kim, Joonkyung and Sim, Joonyeol and Kim, Woojun and Sycara, Katia and Nam, Changjoo},
  journal={arXiv preprint arXiv:2506.03834},
  year={2025}
}

@article{he2024agile,
  title={Agile but safe: Learning collision-free high-speed legged locomotion},
  author={He, Tairan and Zhang, Chong and Xiao, Wenli and He, Guanqi and Liu, Changliu and Shi, Guanya},
  journal={arXiv preprint arXiv:2401.17583},
  year={2024}
}

@inproceedings{xiao2025anycar,
  title={Anycar to anywhere: Learning universal dynamics model for agile and adaptive mobility},
  author={Xiao, Wenli and Xue, Haoru and Tao, Tony and Kalaria, Dvij and Dolan, John M and Shi, Guanya},
  booktitle={2025 IEEE International Conference on Robotics and Automation (ICRA)},
  pages={8819--8825},
  year={2025},
  organization={IEEE}
}

@article{xu2025navrl,
  title={Navrl: Learning safe flight in dynamic environments},
  author={Xu, Zhefan and Han, Xinming and Shen, Haoyu and Jin, Hanyu and Shimada, Kenji},
  journal={IEEE Robotics and Automation Letters},
  year={2025},
  publisher={IEEE}
}

@article{shah2022robotic,
  title={Robotic Navigation with Large Pre-Trained Models of Language},
  author={Shah, Dhruv and Osinski, Blazej and Ichter, Brian and Levine, Sergey},
  journal={Vision, and Action},
  year={2022}
}

@inproceedings{long2024discuss,
  title={Discuss before moving: Visual language navigation via multi-expert discussions},
  author={Long, Yuxing and Li, Xiaoqi and Cai, Wenzhe and Dong, Hao},
  booktitle={2024 IEEE International Conference on Robotics and Automation (ICRA)},
  pages={17380--17387},
  year={2024},
  organization={IEEE}
}

@article{yao2025towards,
  title={Towards Generalizable Safety in Crowd Navigation via Conformal Uncertainty Handling},
  author={Yao, Jianpeng and Zhang, Xiaopan and Xia, Yu and Wang, Zejin and Roy-Chowdhury, Amit K and Li, Jiachen},
  journal={arXiv preprint arXiv:2508.05634},
  year={2025}
}

@article{clevert2015fast,
  title={Fast and accurate deep network learning by exponential linear units (elus)},
  author={Clevert, Djork-Arn{\'e} and Unterthiner, Thomas and Hochreiter, Sepp},
  journal={arXiv preprint arXiv:1511.07289},
  volume={4},
  number={5},
  pages={11},
  year={2015}
}

@article{wang2025omni,
  title={Omni-Perception: Omnidirectional Collision Avoidance for Legged Locomotion in Dynamic Environments},
  author={Wang, Zifan and Ma, Teli and Jia, Yufei and Yang, Xun and Zhou, Jiaming and Ouyang, Wenlong and Zhang, Qiang and Liang, Junwei},
  journal={arXiv preprint arXiv:2505.19214},
  year={2025}
}

@article{shah2023vint,
  title={ViNT: A foundation model for visual navigation},
  author={Shah, Dhruv and Sridhar, Ajay and Dashora, Nitish and Stachowicz, Kyle and Black, Kevin and Hirose, Noriaki and Levine, Sergey},
  journal={arXiv preprint arXiv:2306.14846},
  year={2023}
}

@article{mittal2023orbit,
   author={Mittal, Mayank and Yu, Calvin and Yu, Qinxi and Liu, Jingzhou and Rudin, Nikita and Hoeller, David and Yuan, Jia Lin and Singh, Ritvik and Guo, Yunrong and Mazhar, Hammad and Mandlekar, Ajay and Babich, Buck and State, Gavriel and Hutter, Marco and Garg, Animesh},
   journal={IEEE Robotics and Automation Letters},
   title={Orbit: A Unified Simulation Framework for Interactive Robot Learning Environments},
   year={2023},
   volume={8},
   number={6},
   pages={3740-3747},
   doi={10.1109/LRA.2023.3270034}
}

@article{yang2023iplanner,
  title={iplanner: Imperative path planning},
  author={Yang, Fan and Wang, Chen and Cadena, Cesar and Hutter, Marco},
  journal={arXiv preprint arXiv:2302.11434},
  year={2023}
}

@article{hirose2019deep,
title={Deep visual mpc-policy learning for navigation},
author={Hirose, Noriaki and Xia, Fei and Mart{'\i}n-Mart{'\i}n, Roberto and Sadeghian, Amir and Savarese, Silvio},
journal={IEEE Robotics and Automation Letters},
volume={4},
number={4},
pages={3184--3191},
year={2019},
publisher={IEEE}
}

@inproceedings{meng2020scaling,
  title={Scaling local control to large-scale topological navigation},
  author={Meng, Xiangyun and Ratliff, Nathan and Xiang, Yu and Fox, Dieter},
  booktitle={2020 IEEE International Conference on Robotics and Automation (ICRA)},
  pages={672--678},
  year={2020},
  organization={IEEE}
}

@article{hirose2023sacson,
  title={Sacson: Scalable autonomous control for social navigation},
  author={Hirose, Noriaki and Shah, Dhruv and Sridhar, Ajay and Levine, Sergey},
  journal={IEEE Robotics and Automation Letters},
  volume={9},
  number={1},
  pages={49--56},
  year={2023},
  publisher={IEEE}
}

@article{wang2024grutopia,
  title={Grutopia: Dream general robots in a city at scale},
  author={Wang, Hanqing and Chen, Jiahe and Huang, Wensi and Ben, Qingwei and Wang, Tai and Mi, Boyu and Huang, Tao and Zhao, Siheng and Chen, Yilun and Yang, Sizhe and others},
  journal={arXiv preprint arXiv:2407.10943},
  year={2024}
}

@article{meng2025aim,
  title={Aim My Robot: Precision Local Navigation to Any Object},
  author={Meng, Xiangyun and Yang, Xuning and Jung, Sanghun and Ramos, Fabio and Jujjavarapu, Sri Sadhan and Paul, Sanjoy and Fox, Dieter},
  journal={IEEE Robotics and Automation Letters},
  year={2025},
  publisher={IEEE}
}

@inproceedings{yang2024depth,
  title={Depth anything: Unleashing the power of large-scale unlabeled data},
  author={Yang, Lihe and Kang, Bingyi and Huang, Zilong and Xu, Xiaogang and Feng, Jiashi and Zhao, Hengshuang},
  booktitle={Proceedings of the IEEE/CVF conference on computer vision and pattern recognition},
  pages={10371--10381},
  year={2024}
}

@inproceedings{ehsani2024spoc,
  title={Spoc: Imitating shortest paths in simulation enables effective navigation and manipulation in the real world},
  author={Ehsani, Kiana and Gupta, Tanmay and Hendrix, Rose and Salvador, Jordi and Weihs, Luca and Zeng, Kuo-Hao and Singh, Kunal Pratap and Kim, Yejin and Han, Winson and Herrasti, Alvaro and others},
  booktitle={Proceedings of the IEEE/CVF Conference on Computer Vision and Pattern Recognition},
  pages={16238--16250},
  year={2024}
}

@inproceedings{cai2024bridging,
  title={Bridging zero-shot object navigation and foundation models through pixel-guided navigation skill},
  author={Cai, Wenzhe and Huang, Siyuan and Cheng, Guangran and Long, Yuxing and Gao, Peng and Sun, Changyin and Dong, Hao},
  booktitle={2024 IEEE International Conference on Robotics and Automation (ICRA)},
  pages={5228--5234},
  year={2024},
  organization={IEEE}
}

@article{zhang2025dreamvla,
  title={DreamVLA: a vision-language-action model dreamed with comprehensive world knowledge},
  author={Zhang, Wenyao and Liu, Hongsi and Qi, Zekun and Wang, Yunnan and Yu, Xinqiang and Zhang, Jiazhao and Dong, Runpei and He, Jiawei and Wang, He and Zhang, Zhizheng and others},
  journal={arXiv preprint arXiv:2507.04447},
  year={2025}
}

@article{sun2025view,
  title={View invariant learning for vision-language navigation in continuous environments},
  author={Sun, Josh Qixuan and Xing, Xiaoying and Weng, Huaiyuan and Yeum, Chul Min and Crowley, Mark},
  journal={arXiv preprint arXiv:2507.08831},
  year={2025}
}

@article{qi2025sofar,
  title={Sofar: Language-grounded orientation bridges spatial reasoning and object manipulation},
  author={Qi, Zekun and Zhang, Wenyao and Ding, Yufei and Dong, Runpei and Yu, Xinqiang and Li, Jingwen and Xu, Lingyun and Li, Baoyu and He, Xialin and Fan, Guofan and others},
  journal={arXiv preprint arXiv:2502.13143},
  year={2025}
}

@article{schulman2017proximal,
  title={Proximal policy optimization algorithms},
  author={Schulman, John and Wolski, Filip and Dhariwal, Prafulla and Radford, Alec and Klimov, Oleg},
  journal={arXiv preprint arXiv:1707.06347},
  year={2017}
}

@article{frantar2022gptq,
  title={Gptq: Accurate post-training quantization for generative pre-trained transformers},
  author={Frantar, Elias and Ashkboos, Saleh and Hoefler, Torsten and Alistarh, Dan},
  journal={arXiv preprint arXiv:2210.17323},
  year={2022}
}

@article{lin2024awq,
  title={Awq: Activation-aware weight quantization for on-device llm compression and acceleration},
  author={Lin, Ji and Tang, Jiaming and Tang, Haotian and Yang, Shang and Chen, Wei-Ming and Wang, Wei-Chen and Xiao, Guangxuan and Dang, Xingyu and Gan, Chuang and Han, Song},
  journal={Proceedings of machine learning and systems},
  volume={6},
  pages={87--100},
  year={2024}
}

@article{wang2025internvla,
  title = {InternVLA-N1: An Open Dual-System Vision-Language Navigation Foundation Model with Learned Latent Plans},
  author = {{Intern Robotics}},
  booktitle = {Arxiv},
  year = {2025},
}

@article{peng2025logoplanner,
  title={Logoplanner: Localization grounded navigation policy with metric-aware visual geometry},
  author={Peng, Jiaqi and Cai, Wenzhe and Yang, Yuqiang and Wang, Tai and Shen, Yuan and Pang, Jiangmiao},
  journal={arXiv preprint arXiv:2512.19629},
  year={2025}
}

@article{zhang2026self,
  title={Self-Imitated Diffusion Policy for Efficient and Robust Visual Navigation},
  author={Zhang, Runhua and Hou, Junyi and Cheng, Changxu and Chen, Qiyi and Wang, Tao and Zhao, Wuyue},
  journal={arXiv preprint arXiv:2601.22965},
  year={2026}
}

@article{zhong2025internscenes,
  title={Internscenes: A large-scale simulatable indoor scene dataset with realistic layouts},
  author={Zhong, Weipeng and Cao, Peizhou and Jin, Yichen and Luo, Li and Cai, Wenzhe and Lin, Jingli and Wang, Hanqing and Lyu, Zhaoyang and Wang, Tai and Dai, Bo and others},
  journal={arXiv preprint arXiv:2509.10813},
  year={2025}
}

@article{huang2026ticvla,
  title={TIC-VLA: A Think-in-Control Vision-Language-Action Model for Robot Navigation in Dynamic Environments},
  author={Zhiyu Huang and Yun Zhang and Johnson Liu and Rui Song and Chen Tang and Jiaqi Ma},
  year={2026},
  eprint={2602.02459},
}

@article{shi2025fastsmartway,
  title={Fast-SmartWay: Panoramic-Free End-to-End Zero-Shot Vision-and-Language Navigation},
  author={Xiangyu Shi and Zerui Li and Yanyuan Qiao and Qi Wu},
  year={2025},
  eprint={2511.00933},
}

@article{castro2025vamos,
  title={VAMOS: A Hierarchical Vision-Language-Action Model for Capability-Modulated and Steerable Navigation},
  author={Mateo Guaman Castro and Sidharth Rajagopal and Daniel Gorbatov and Matt Schmittle and Rohan Baijal and Octi Zhang and Rosario Scalise and Sidharth Talia and Emma Romig and Celso de Melo and Byron Boots and Abhishek Gupta},
  year={2025},
  eprint={2510.20818},
}

@article{yu2025correctnav,
  title={CorrectNav: Self-Correction Flywheel Empowers Vision-Language-Action Navigation Model},
  author={Zhuoyuan Yu and Yuxing Long and Zihan Yang and Chengyan Zeng and Hongwei Fan and Jiyao Zhang and Hao Dong},
  year={2025},
  eprint={2508.10416},
}

@inproceedings{cheng2025navila,
  author    = {An-Chieh Cheng and Yandong Ji and Zhaojing Yang and Zaitian Gongye and Xueyan Zou and Jan Kautz and Erdem Biyik and Hongxu Yin and Sifei Liu and Xiaolong Wang},
  title     = {NaVILA: Legged Robot Vision-Language-Action Model for Navigation},
  booktitle = {Proceedings of Robotics: Science and Systems},
  year      = {2025},
  address   = {Los Angeles, CA, USA},
  month     = jun
}

@inproceedings{liu2025citywalker,
  title={Citywalker: Learning embodied urban navigation from web-scale videos},
  author={Liu, Xinhao and Li, Jintong and Jiang, Yicheng and Sujay, Niranjan and Yang, Zhicheng and Zhang, Juexiao and Abanes, John and Zhang, Jing and Feng, Chen},
  booktitle={Proceedings of the Computer Vision and Pattern Recognition Conference},
  pages={6875--6885},
  year={2025}
}

@article{hirose2025driveanywhere,
  title   = {Learning to Drive Anywhere with Model-Based Reannotation},
  author  = {Hirose, Noriaki and Ignatova, Lydia and Stachowicz, Kyle and Glossop, Catherine and Levine, Sergey and Shah, Dhruv},
  journal = {arXiv preprint arXiv:2505.05592},
  year    = {2025},
  url     = {https://arxiv.org/abs/2505.05592}
}

@article{hirose2025omnivla,
  title   = {OmniVLA: An Omni-Modal Vision-Language-Action Model for Robot Navigation},
  author  = {Hirose, Noriaki and Glossop, Catherine and Shah, Dhruv and Levine, Sergey},
  journal = {arXiv preprint arXiv:2509.19480},
  year    = {2025},
  url     = {https://arxiv.org/abs/2509.19480}
}

@article{zhang2025navfom,
  title   = {Embodied Navigation Foundation Model},
  author  = {Zhang, Jiazhao and Li, Anqi and Qi, Yunpeng and Li, Minghan and Liu, Jiahang and Wang, Shaoan and Liu, Haoran and Zhou, Gengze and Wu, Yuze and Li, Xingxing and Fan, Yuxin and Li, Wenjun and Chen, Zhibo and Gao, Fei and Wu, Qi and Zhang, Zhizheng and Wang, He},
  journal = {arXiv preprint arXiv:2509.12129},
  year    = {2025},
  url     = {https://arxiv.org/abs/2509.12129}
}

@article{yang2025lohovla,
  title = {LoHoVLA: A Unified Vision-Language-Action Model for Long-Horizon Embodied Tasks},
  author = {Yi Yang and Jiaxuan Sun and Siqi Kou and Yihan Wang and Zhijie Deng},
  booktitle = {Arxiv},
  year = {2025},
}

@article{xue2026omninav,
  title = {OmniNav: A Unified Framework for Prospective Exploration and Visual-Language Navigation},
  author = {Xinda Xue and Junjun Hu and Minghua Luo and Shichao Xie and Jintao Chen and Zixun Xie and Kuichen Quan and Wei Guo and Mu Xu and Zedong Chu},
  booktitle = {Arxiv},
  year = {2026},
}

@article{wang2026imaginenavpp,
  title = {ImagineNav++: Prompting Vision-Language Models as Embodied Navigator through Scene Imagination},
  author = {Teng Wang and Xinxin Zhao and Wenzhe Cai and Changyin Sun},
  booktitle = {Arxiv},
  year = {2026},
}

@article{zhang2025mem2ego,
  title = {Mem2Ego: Empowering Vision-Language Models with Global-to-Ego Memory for Long-Horizon Embodied Navigation},
  author = {Lingfeng Zhang and Yuecheng Liu and Zhanguang Zhang and Matin Aghaei and Yaochen Hu and Hongjian Gu and Mohammad Ali Alomrani and David Gamaliel Arcos Bravo and Raika Karimi and Atia Hamidizadeh and Haoping Xu and Guowei Huang and Zhanpeng Zhang and Tongtong Cao and Weichao Qiu and Xingyue Quan and Jianye Hao and Yuzheng Zhuang and Yingxue Zhang},
  booktitle = {Arxiv},
  year = {2025},
}

@article{lee2026rvnbench,
  title = {RVN-Bench: A Benchmark for Reactive Visual Navigation},
  author = {Jaewon Lee and Jaeseok Heo and Gunmin Lee and Howoong Jun and Jeongwoo Oh and Songhwai Oh},
  booktitle = {Arxiv},
  year = {2026},
}

@InProceedings{pmlr-v270-hirose25b,
  title = 	 {LeLaN: Learning A Language-Conditioned Navigation Policy from In-the-Wild Video},
  author =       {Hirose, Noriaki and Glossop, Catherine and Sridhar, Ajay and Mees, Oier and Levine, Sergey},
  booktitle = 	 {Proceedings of The 8th Conference on Robot Learning},
  pages = 	 {666--688},
  year = 	 {2025},
  editor = 	 {Agrawal, Pulkit and Kroemer, Oliver and Burgard, Wolfram},
  volume = 	 {270},
  series = 	 {Proceedings of Machine Learning Research},
  month = 	 {06--09 Nov},
  publisher =    {PMLR},
  url = 	 {https://proceedings.mlr.press/v270/hirose25b.html}
}

@inproceedings{he2026seeing,
  title={From Seeing to Experiencing: Scaling Navigation Foundation Models with Reinforcement Learning},
  author={Honglin He and Yukai Ma and Brad Squicciarini and Wayne Wu and Bolei Zhou},
  booktitle={International Conference on Learning Representations},
  year={2026},
  note={Poster},
  url={https://openreview.net/forum?id=0c7nAZjyr5}
}

@inproceedings{roth2025fdm,
  title={Learned Perceptive Forward Dynamics Model for Safe and Platform-aware Robotic Navigation},
  author={Roth, Pascal and Frey, Jonas and Cadena, Cesar and Hutter, Marco},
  booktitle={Robotics: Science and Systems (RSS)},
  year={2025}
}

@article{zhu2026hicrowd,
  title={HiCrowd: Hierarchical Crowd Flow Alignment for Dense Human Environments},
  author={Zhu, Yufei and Yang, Shih-Min and Magnusson, Martin and Wang, Allan},
  journal={arXiv preprint arXiv:2602.05608},
  year={2026}
}

@inproceedings{seneviratne2025halo,
  title={HALO: Human Preference Aligned Offline Reward Learning for Robot Navigation},
  author={Seneviratne, Gershom and An, Jianyu and Ellahy, Sahire and Weerakoon, Kasun and Elnoor, Mohamed Bashir and Kannan, Jonathan Deepak and Sunil, Amogha Thalihalla and Manocha, Dinesh},
  booktitle={Conference on Robot Learning (CoRL)},
  year={2025}
}

@InProceedings{Wang_2025_ICCV,
  author    = {Wang, Liuyi and Xia, Xinyuan and Zhao, Hui and Wang, Hanqing and Wang, Tai and Chen, Yilun and Liu, Chengju and Chen, Qijun and Pang, Jiangmiao},
  title     = {Rethinking the Embodied Gap in Vision-and-Language Navigation: A Holistic Study of Physical and Visual Disparities},
  booktitle = {Proceedings of the IEEE/CVF International Conference on Computer Vision (ICCV)},
  month     = {October},
  year      = {2025},
  pages     = {9455-9465},
  url       = {https://openaccess.thecvf.com/content/ICCV2025/papers/Wang_Rethinking_the_Embodied_Gap_in_Vision-and-Language_Navigation_A_Holistic_Study_ICCV_2025_paper.pdf}
}

@InProceedings{Zhong_2025_ICCV,
  author    = {Zhong, Yufeng and Feng, Chengjian and Yan, Feng and Liu, Fanfan and Zheng, Liming and Ma, Lin},
  title     = {RoboTron-Nav: A Unified Framework for Embodied Navigation Integrating Perception, Planning, and Prediction},
  booktitle = {Proceedings of the IEEE/CVF International Conference on Computer Vision (ICCV)},
  month     = {October},
  year      = {2025},
  pages     = {6416-6425},
  url       = {https://openaccess.thecvf.com/content/ICCV2025/papers/Zhong_RoboTrom-Nav_A_Unified_Framework_for_Embodied_Navigation_Integrating_Perception_Planning_ICCV_2025_paper.pdf}
}

@InProceedings{Xu_2025_ICCV,
  author    = {Xu, Peiran and Gong, Xicheng and Mu, Yadong},
  title     = {NavQ: Learning a Q-Model for Foresighted Vision-and-Language Navigation},
  booktitle = {Proceedings of the IEEE/CVF International Conference on Computer Vision (ICCV)},
  month     = {October},
  year      = {2025},
  pages     = {6327-6341},
  url       = {https://openaccess.thecvf.com/content/ICCV2025/papers/Xu_NavQ_Learning_a_Q-Model_for_Foresighted_Vision-and-Language_Navigation_ICCV_2025_paper.pdf}
}

@InProceedings{Qin_2025_ICCV,
  author    = {Qin, Liang and Wang, Min and Li, Peiwei and Zhou, Wengang and Li, Houqiang},
  title     = {Active Perception Meets Rule-Guided RL: A Two-Phase Approach for Precise Object Navigation in Complex Environments},
  booktitle = {Proceedings of the IEEE/CVF International Conference on Computer Vision (ICCV)},
  month     = {October},
  year      = {2025},
  pages     = {7603-7612},
  url       = {https://openaccess.thecvf.com/content/ICCV2025/papers/Qin_Active_Perception_Meets_Rule-Guided_RL_A_Two-Phase_Approach_for_Precise_ICCV_2025_paper.pdf}
}

@inproceedings{chhablani2025embodiedsplat,
  title={Embodiedsplat: Personalized real-to-sim-to-real navigation with gaussian splats from a mobile device},
  author={Chhablani, Gunjan and Ye, Xiaomeng and Irshad, Muhammad Zubair and Kira, Zsolt},
  booktitle={Proceedings of the IEEE/CVF International Conference on Computer Vision},
  pages={25431--25441},
  year={2025}
}

@inproceedings{pan2025lookout,
  title={Lookout: Real-world humanoid egocentric navigation},
  author={Pan, Boxiao and Harley, Adam W and Engelmann, Francis and Liu, C Karen and Guibas, Leonidas J},
  booktitle={Proceedings of the IEEE/CVF International Conference on Computer Vision},
  pages={24977--24988},
  year={2025}
}

@inproceedings{qiao2024llm,
  title={LLM as copilot for coarse-grained vision-and-language navigation},
  author={Qiao, Yanyuan and Liu, Qianyi and Liu, Jiajun and Liu, Jing and Wu, Qi},
  booktitle={European Conference on Computer Vision},
  pages={459--476},
  year={2024},
  organization={Springer}
}

@inproceedings{lu2024pret,
  title={Pret: Planning with directed fidelity trajectory for vision and language navigation},
  author={Lu, Renjie and Meng, Jingke and Zheng, Wei-Shi},
  booktitle={European Conference on Computer Vision},
  pages={72--88},
  year={2024},
  organization={Springer}
}

@inproceedings{xu2024disco,
  title={DISCO: Embodied Navigation and Interaction via Differentiable Scene Semantics and Dual-level Control},
  author={Xu, Xinyu and Luo, Shengcheng and Yang, Yanchao and Li, Yong-Lu and Lu, Cewu},
  booktitle={European Conference on Computer Vision},
  pages={108--125},
  year={2024},
  organization={Springer}
}

@inproceedings{cui2024frontier,
  title={Frontier-enhanced topological memory with improved exploration awareness for embodied visual navigation},
  author={Cui, Xinru and Liu, Qiming and Liu, Zhe and Wang, Hesheng},
  booktitle={European Conference on Computer Vision},
  pages={296--313},
  year={2024},
  organization={Springer}
}

\clearpage
\section{Appendix}

This section supplements the RL expert description in the main paper by providing additional details on the expert environments, the RL formulation, the training setup, and the corresponding visualizations. In the main paper, the RL experts serve as capability-specific teachers for the student VLA model. Accordingly, this appendix focuses on how the three experts are constructed and trained.

\subsection{RL Expert Environments}

\begin{figure}[t]
\centering
\includegraphics[width=\linewidth]{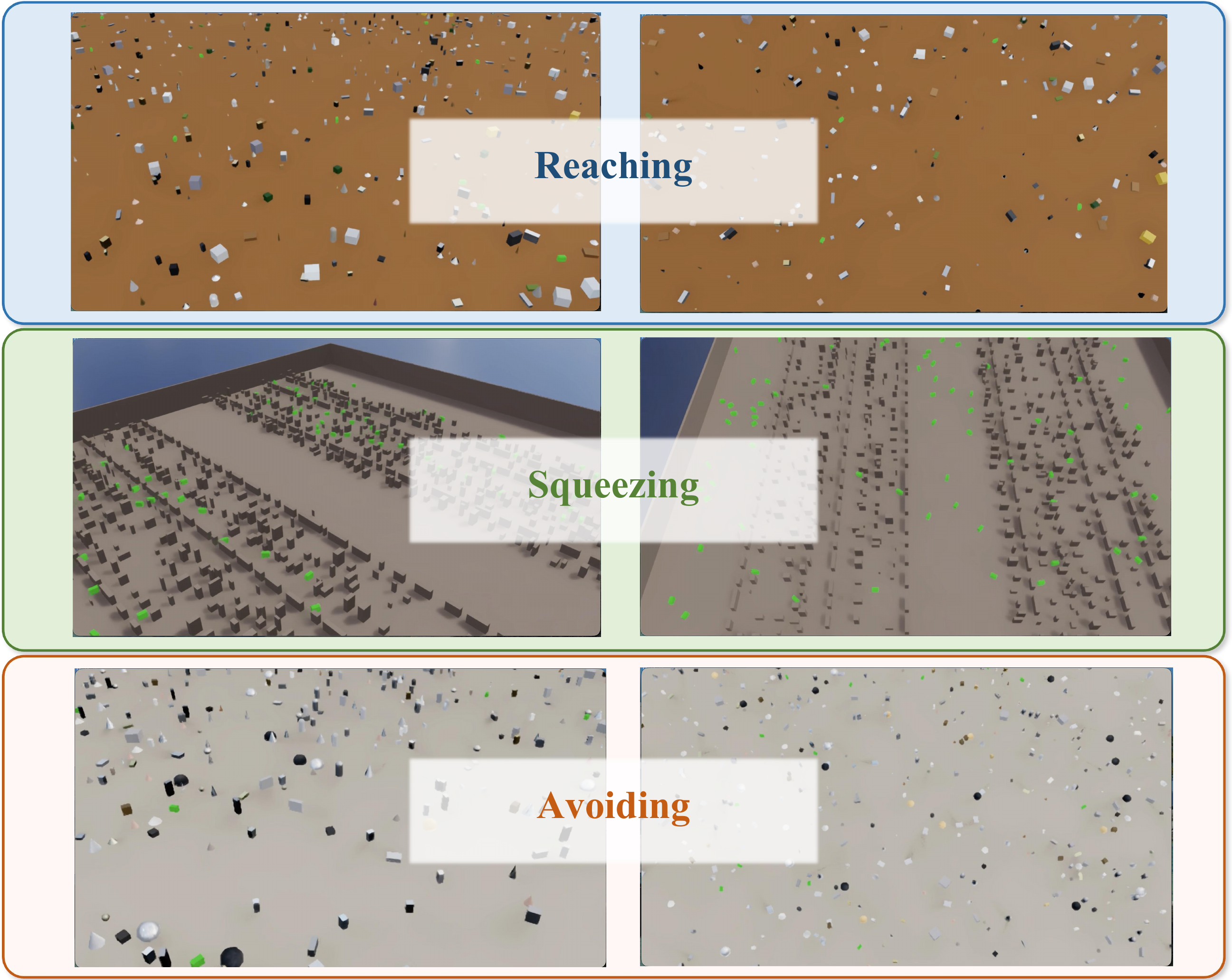}
\caption{Representative views of the three RL expert environments. From top to bottom: reaching, squeezing, and avoiding.}
\label{fig:rl_expert_environments}
\end{figure}

We train three RL experts, each specializing in one navigation capability: reaching, squeezing, and avoiding. All three experts operate in IsaacLab simulation and control the same omnidirectional robot abstraction. Their differences arise from the environment geometry and obstacle dynamics. The visualization results are shown in \Cref{fig:rl_expert_environments}.

\subsubsection{Reaching Environment}

The reaching environment is designed to teach robust goal-directed navigation in large-scale cluttered scenes with static obstacles. The robot is required to approach and reach a designated point goal while avoiding collisions with surrounding objects. In the released configuration, all parallel environments share a $270~\mathrm{m} \times 270~\mathrm{m}$ terrain template, while each environment instance instantiates $65$ obstacle actors, covering randomized cuboids, cones, capsules, low objects, floating cuboids, and long sticks. This corresponds to an average obstacle density of approximately $8.9 \times 10^{-4}$ actors per square meter with respect to the terrain footprint of a single environment instance. Under the setting of $128$ parallel environments, the simulator therefore contains $8{,}320$ obstacle instances in total.

The robot is reset within a central $220~\mathrm{m} \times 220~\mathrm{m}$ region, while the point goal is sampled with distances of up to $30$ meters. Since the obstacles remain static, the expert does not need to anticipate future obstacle motion. Instead, it must learn long-range, collision-free navigation through geometrically complex clutter while maintaining steady progress toward the goal. This environment is therefore intended to train the reaching capability in isolation.

\subsubsection{Squeezing Environment}

The squeezing environment is designed to teach navigation through narrow free-space corridors. In the released configuration, all parallel environments share a $70~\mathrm{m} \times 70~\mathrm{m}$ terrain template generated by a dense pillar terrain generator. The underlying heightfield is first partitioned into a $12 \times 12$ array of local regions. Within each region, the generator places short rectangular pillars in a staggered lattice with nominal spacing of approximately $1.9~\mathrm{m}$, random widths of $0.4$--$0.8~\mathrm{m}$, and random lengths of $0.3$--$1.2~\mathrm{m}$, together with small random offsets and occasional omissions so that the free space forms irregular gaps rather than a perfectly regular grid. In addition, the terrain keeps a central $3.0~\mathrm{m} \times 3.0~\mathrm{m}$ flat platform, clears three wider flat longitudinal bands for the left spawn side, the middle crossing corridor, and the right spawn side, and inserts three thin wall barriers with random door openings on the left and right obstacle fields. To reduce overfitting to a single layout, the procedural terrain is periodically regenerated during training, so the pillar arrangement, wall openings, and traversable passages change over time. As a result, the environment forms repeated narrow but traversable side-to-center passages instead of a uniform pillar field.

The robot is reset from one of two side strips, with $x \in [-33,-28]$ or $x \in [28,33]$ and $y \in [-33,33]$, while goals are sampled from a corridor-centered command range with $x \in [-5,5]$ and $y \in [-30,30]$. Moreover, the command generator operates in a corridor mode that usually preserves the robot's current $y$ coordinate and therefore emphasizes side-to-side traversal through narrow gaps, while occasionally resampling $y$ to diversify the path geometry. Compared with reaching, the squeezing scene emphasizes lateral clearance reasoning. The side views are especially important in this setting because the robot must determine whether the available gap is sufficiently wide for traversal.

\subsubsection{Avoiding Environment}

The avoiding environment is designed to teach active collision avoidance in dynamic scenes. In the released configuration, all parallel environments share a $150~\mathrm{m} \times 150~\mathrm{m}$ terrain template, while each environment instance instantiates $40$ moving obstacles, including diverse cuboids, cones, low objects, floating objects, and spheres. This corresponds to an average obstacle density of approximately $1.8 \times 10^{-3}$ actors per square meter with respect to the terrain footprint of a single environment instance. Under the setting of $128$ parallel environments, the simulator therefore contains $5{,}120$ dynamic obstacle instances in total.

The robot-goal distance is sampled up to $10$ meters. Each dynamic obstacle resamples a motion direction uniformly over $360^\circ$ and a speed uniformly from $0.5$ to $1.5~\mathrm{m/s}$, with the motion command resampled every $1$--$15$ seconds and clipped to remain inside the scene bounds. The expert must therefore maintain progress toward the goal while repeatedly adjusting its trajectory to avoid nearby moving objects. This environment is therefore intended to train the avoiding capability under dynamic interaction.

\subsection{RL Formulation}

Unlike the student VLA policy, which consumes multi-view RGB observations, the RL experts operate on multi-view depth observations and predict omnidirectional velocity commands.
\subsubsection{Task Definition}

At time step $t$, the relative point goal is denoted by $g_t = [g_t^x, g_t^y]$, the RL observation by $O_t^{RL}$, and the velocity control policy by $\pi$. The policy output is denoted by $a_t = [a_t^x, a_t^y, a_t^{yaw}]$, whose three components correspond to planar motion along the $x$ axis, planar motion along the $y$ axis, and yaw control, respectively. The RL task is to learn

$$
\pi(O_t^{RL}, g_t) \mapsto a_t.
$$

The objective is to reach the designated goal with collision-free and efficient motion.

\subsubsection{Model Architecture}

The four depth observations at time step $t$ are denoted by $d_t^{front}$, $d_t^{right}$, $d_t^{back}$, and $d_t^{left}$, and the previous action is denoted by $a_{t-1}$. The RL input is

$$
O_t^{RL} = [d_t^{front}, d_t^{right}, d_t^{back}, d_t^{left}, a_{t-1}, g_t].
$$

Each depth view is encoded by a ResNet-18 backbone. Denoting the encoded features by $f_t^{front}$, $f_t^{right}$, $f_t^{back}$, and $f_t^{left}$, the history token from the previous step by $h_{t-1}$, and the fused policy feature by $z_t$, the model computes

$$
z_t = [f_t^{front}, f_t^{right}, f_t^{back}, f_t^{left}, a_{t-1}, g_t, h_{t-1}].
$$

The fused feature $z_t$ is processed by a history-aware actor-critic with three-layer MLP heads. Here, $h_{t-1}$ is taken from the last hidden layer at the previous step and provides lightweight temporal memory without an explicit recurrent module.

\subsubsection{Action Parameterization}

The normalized policy output is $a_t = [a_t^x, a_t^y, a_t^{yaw}]$, and the executed omnidirectional velocity command is $v_t = [v_t^x, v_t^y, v_t^{yaw}]$. The velocity limits are denoted by $v_{max} = [v_{max}^x, v_{max}^y, v_{max}^{yaw}]$, where $v_{max}^x = 1.5~\mathrm{m/s}$, $v_{max}^y = 1.0~\mathrm{m/s}$, and $v_{max}^{yaw} = \pi / 4.0~\mathrm{rad/s}$. The executed command is obtained by element-wise clipping and scaling:

$$
v_t = \max\{\min\{a_t, 1.0\}, -1.0\} \times v_{max}.
$$

This parameterization yields continuous omnidirectional planar control.

\subsubsection{Reward and Termination}

Following the main paper, the per-step reward for capability $Cap. \in \{\text{reaching}, \text{squeezing}, \text{avoiding}\}$ is written as $r_{Cap.}$, with grouped components $r_{goal}$, $r_{step}$, $r_{reg}$, and $r_{col}$ weighted by $\alpha_{Cap.}$, $\beta_{Cap.}$, $\gamma_{Cap.}$, and $\delta_{Cap.}$, respectively:

\begin{equation}
r_{Cap.} = \alpha_{Cap.} r_{goal} + \beta_{Cap.} r_{step} + \gamma_{Cap.} r_{reg} + \delta_{Cap.} r_{col}.
\end{equation}

We denote the goal distance by $p_t = \lVert g_t \rVert_2$, the stepwise progress by $\Delta p_t = p_{t-1} - p_t$, the planar velocity by $v_t^{xy} = [v_t^x, v_t^y]$, and the yaw rate by $\omega_t = v_t^{yaw}$. We further use $\bar{\omega}_t = \min(|\omega_t|, 0.4)$ for the clipped yaw magnitude, $b_t = \max(0, -v_t^x)$ for the backward-motion magnitude, and $\Delta a_t = |a_t - a_{t-1}|$ for the action variation. The success thresholds are denoted by $\epsilon_d$, $\epsilon_v$, and $\epsilon_\omega$, the near-goal thresholds by $\tau_d$ and $\tau_\omega$, and the failure indicators by $c_t^{contact}$ and $c_t^{pose}$. For compactness, we define
\begin{equation}
\begin{aligned}
s_t &= \lambda_{s}\mathbb{I}[p_t < \epsilon_d \land \lVert v_t^{xy} \rVert_2 < \epsilon_v \land |\omega_t| < \epsilon_\omega], \\
m_t &= \mathbb{I}[\text{the episode remains active at step } t], \\
n_t &= \mathbb{I}[p_t < \tau_d \land |\omega_t| < \tau_\omega].
\end{aligned}
\end{equation}
where $\lambda_{s}$ is a weight in $s_t$ equation.
The forward-motion heading penalty $\theta_t$ is defined as
\begin{equation}
\theta_t =
\mathbb{I}\big[v_t^x > 0 \land (|v_t^x| > 0.1 \lor |v_t^y| > 0.1)\big]
\cdot \left|\arctan2(v_t^y, v_t^x)\right|.
\end{equation}

The grouped terms are instantiated as
\begin{equation}
\begin{aligned}
r_{goal} &= s_t + \Delta p_t, \\
r_{step} &= m_t, \\
r_{reg} &= n_t \left(\exp(-\lVert v_t^{xy} \rVert_2) + \exp(-|\omega_t|)\right)
- \bar{\omega}_t - b_t - \lVert \Delta a_t \rVert_2^2 - \theta_t, \\
r_{col} &= c_t^{contact} + c_t^{pose}.
\end{aligned}
\end{equation}

This decomposition matches the semantics of \Cref{eq3}: $r_{goal}$ combines dense progress and terminal success, $r_{step}$ induces a constant time penalty, $r_{reg}$ collects motion regularizers, and $r_{col}$ penalizes collisions and failure states. The capability-specific goal-reaching thresholds and reward coefficients are summarized in Table~\ref{tab:rl_capability_hparams}. The squeezing expert uses a stronger goal term, while the avoiding expert applies the weakest regularization.

\begin{table}[t]

\centering
\small
\setlength{\tabcolsep}{5pt}
\caption{Capability-specific goal-reaching thresholds and reward coefficients.}
\label{tab:rl_capability_hparams}
\begin{tabular}{lcccccccc}
\hline
Expert & $\epsilon_d$ & $\epsilon_v$ & $\epsilon_\omega$ & $\alpha_{Cap.}$ & $\beta_{Cap.}$ & $\gamma_{Cap.}$ & $\delta_{Cap.}$ & $\lambda_{s}$ \\
 & (m) & (m/s) & (rad/s) &  &  &  &  &  \\
\hline
Reaching  & $0.5$ & $0.3$ & $0.1$ & $1.2$ & $-0.05$ & $0.05$ & $-15$ & $30$ \\
Squeezing & $0.5$ & $0.3$ & $0.1$ & $1.5$ & $-0.05$ & $0.02$ & $-15$ & $30$ \\
Avoiding  & $0.5$ & $0.3$ & $0.1$ & $1.2$ & $-0.05$ & $0$    & $-15$ & $25$ \\
\hline
\end{tabular}
\end{table}

Let $T_{max}$ denote the maximum episode length of the corresponding scene. The termination indicator $\mathcal{T}_t$ is defined as

\begin{equation}
\mathcal{T}_t = \mathbb{I}[p_t < \epsilon_d \land \lVert v_t^{xy} \rVert_2 < \epsilon_v \land |\omega_t| < \epsilon_\omega] \lor c_t^{contact} \lor c_t^{pose} \lor \mathbb{I}[t > T_{max}],
\end{equation}

which terminates an episode upon successful goal reaching, failure, or timeout.

\subsection{RL Training Details}

All three RL experts are trained in IsaacLab with a shared policy and optimization setup. Beyond the core settings reported in the main paper, the released configurations also specify common rollout and PPO hyperparameters, which we summarize here for completeness.

\subsubsection{Policy and Optimization}

All three experts are trained with the Proximal Policy Optimization (PPO) algorithm. The policy adopts a history-aware actor-critic architecture. Both the actor and the critic are implemented as three-layer MLPs with hidden dimensions $[512, 256, 128]$ and use the ELU activation function. The action distribution is initialized with a standard deviation of $0.2$ to encourage early exploration.

\subsubsection{Simulation and Training Setup}

Each RL expert is trained in IsaacLab using $N = 128$ parallel environments. Training each expert takes approximately $8$--$12$ hours on one NVIDIA RTX 4090 GPU. The robot is abstracted as a cuboid of size $[0.70~\mathrm{m}, 0.35~\mathrm{m}, 0.50~\mathrm{m}]$ in order to improve physical simulation efficiency and rendering speed. The depth observations are clipped to the range $[0.01~\mathrm{m}, 4.0~\mathrm{m}]$ to suppress noisy or invalid measurements. During training, Gaussian noise is added to the proprioceptive velocity observations to improve robustness. In the released configurations, the simulation step is fixed to $0.2$ s with decimation $1$, and the episode horizons are set to $60$ s for reaching and avoiding, $45$ s for squeezing. More detailed settings are provided in \Cref{tab:rl_expert_training_setup} and \Cref{tab:rl_ppo_hparams}.

\begin{table}[t]
\centering
\caption{RL expert training setup.}
\label{tab:rl_expert_training_setup}
\begin{tabular}{ll}
\hline
Item & Value \\
\hline
RL algorithm & PPO \\
Parallel environments & 128 \\
Actor hidden dimensions & $[512, 256, 128]$ \\
Critic hidden dimensions & $[512, 256, 128]$ \\
Activation & ELU \\
Initial action noise std & 0.2 \\
Depth clipping range & $[0.01~\mathrm{m}, 4.0~\mathrm{m}]$ \\
Robot abstraction size & $[0.70~\mathrm{m}, 0.35~\mathrm{m}, 0.50~\mathrm{m}]$ \\
Training hardware & one NVIDIA RTX 4090 \\
Training time per expert & 8--12 hours \\
\hline
\end{tabular}
\end{table}

\begin{table}[h]
\centering
\small
\setlength{\tabcolsep}{5pt}
\caption{Additional rollout and PPO hyperparameters from the released RL configurations. These settings are shared across the three experts unless noted otherwise.}
\label{tab:rl_ppo_hparams}
\begin{tabular}{ll}
\hline
Setting & Value \\
\hline
Rollout steps per environment & 128 \\
PPO learning epochs & 8 \\
PPO mini-batches & 16 \\
PPO clip parameter & 0.3 \\
Value loss coefficient & 1.0 \\
Entropy coefficient & 0.01 \\
Learning rate & $1\times 10^{-6}$ \\
Learning-rate schedule & adaptive \\
Discount factor $\gamma$ & 0.99 \\
GAE parameter $\lambda$ & 0.95 \\
Target KL & 0.02 \\
Max gradient norm & 1.0 \\
Simulation step & $0.2$ s \\
Control decimation & 1 \\
Episode horizon (reaching / squeezing / avoiding) & 60 / 45 / 60 s \\
Runner iterations & 25001 \\
\hline
\end{tabular}
\end{table}


\subsection{Additional Training and Visualization Evidence}

\begin{figure}[t]
\centering
\includegraphics[width=\linewidth]{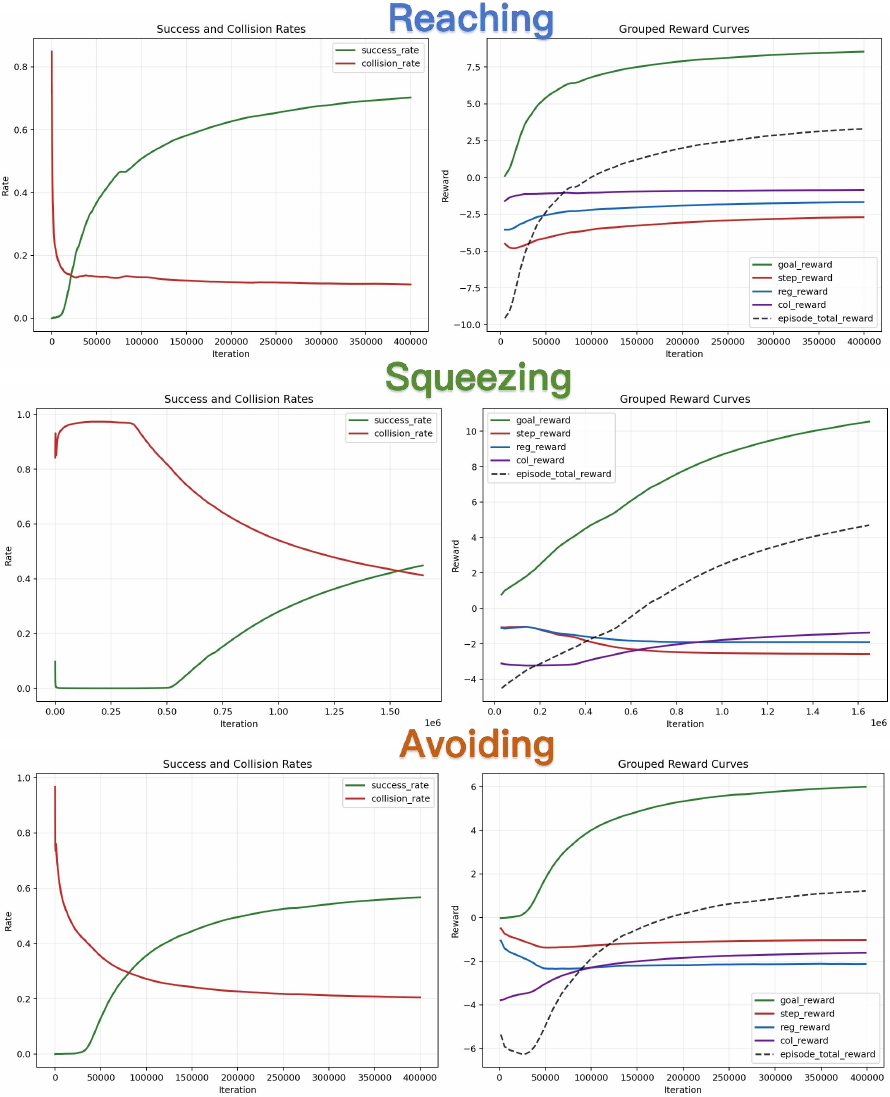}
\caption{Training curves of the three RL experts. }
\label{fig:rl_training_curves}
\end{figure}

\begin{figure}[h]
  \centering
  \includegraphics[width=1.\textwidth]{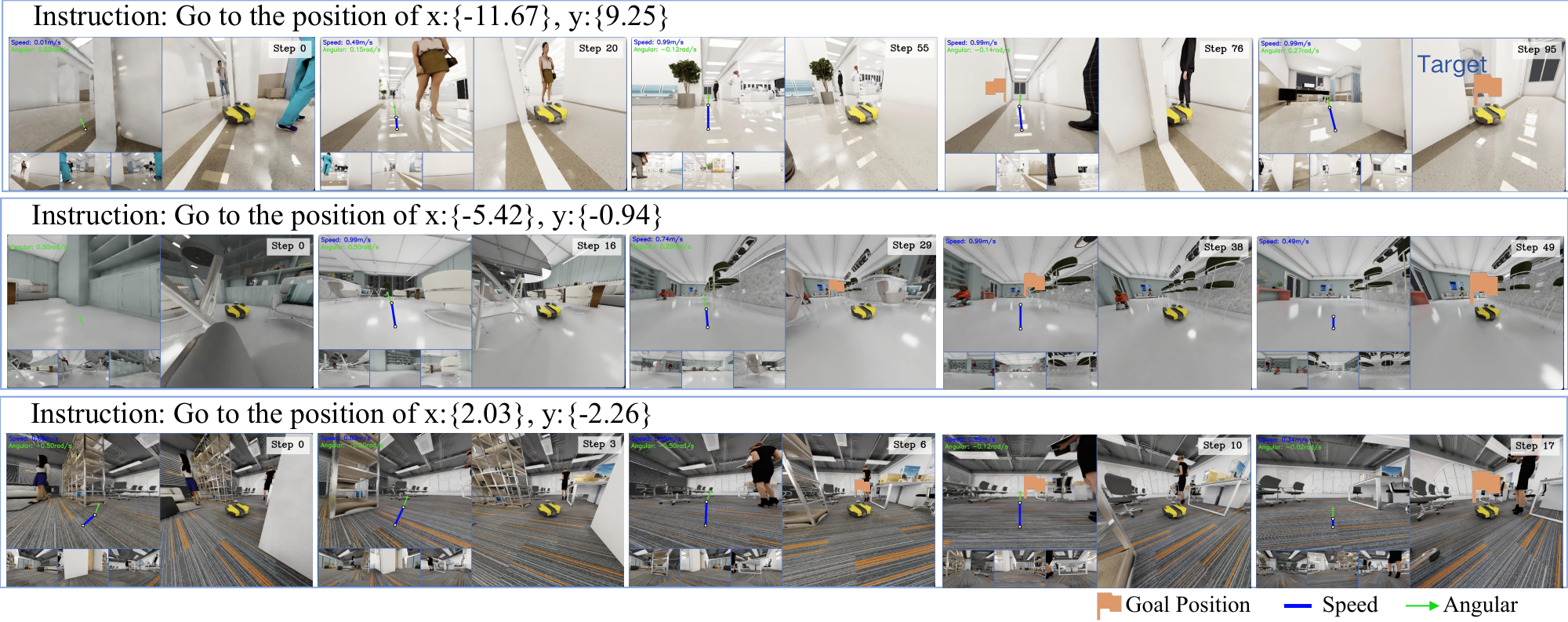}
  \caption{
    Experimental Visualization of MM-Nav on InternVLA-N1 System-1 point-goal navigation benchmark.
  }
  \label{fig:navdp-ben}
  
\end{figure}

\begin{figure}[h]
  \centering
  \includegraphics[width=1.\textwidth]{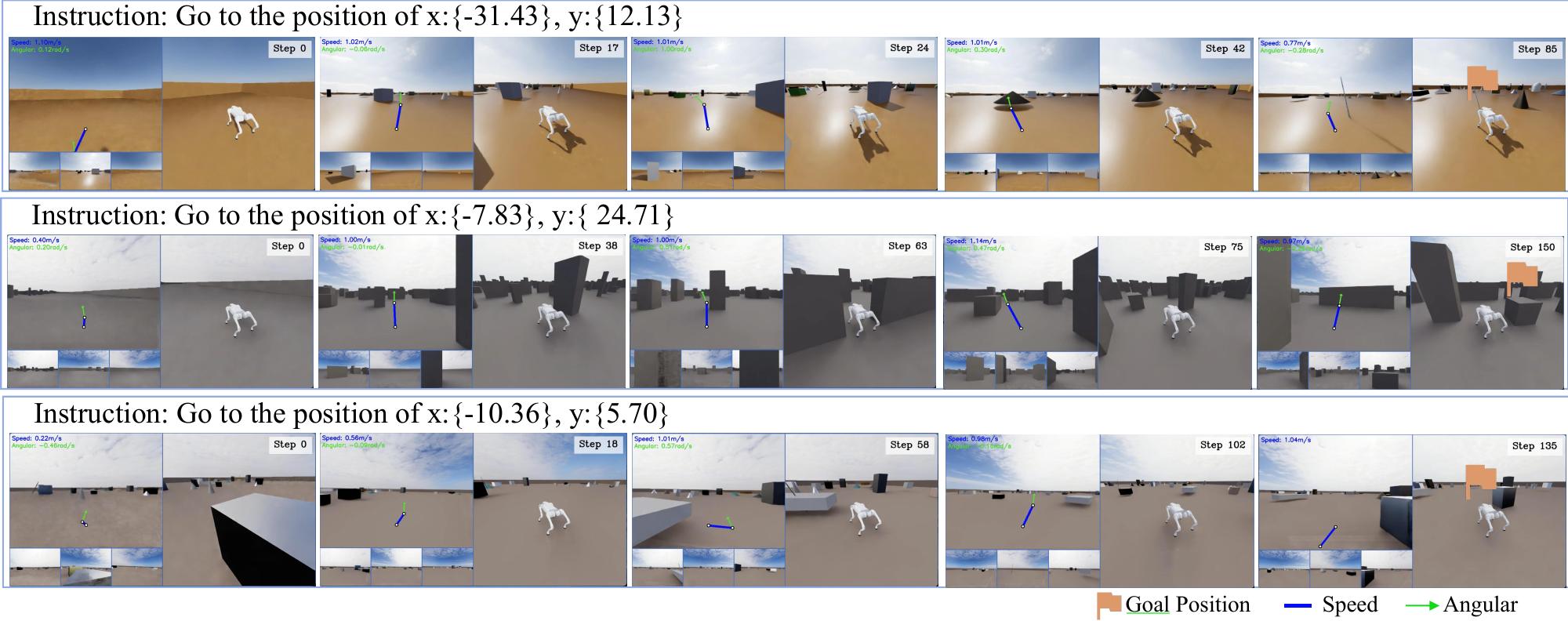}
  \caption{
    Experimental Visualization of MM-Nav on MM-Nav point-goal navigation benchmark.
  }
  \label{fig:mmnav-ben}
  
\end{figure}

Figure~\ref{fig:rl_training_curves} reports the training curves of the three RL experts. As shown in the figure, all three experts exhibit stable optimization trends and converge within the training budget despite their different scene difficulty and interaction structure. For qualitative reference, Figure~\ref{fig:navdp-ben} presents example rollouts on the InternVLA-N1 System-1 point-goal navigation benchmark, while Figure~\ref{fig:mmnav-ben} shows representative trajectories on the MM-Nav point-goal navigation benchmark. Together, these figures complement the quantitative training curves with visual evidence of the learned navigation behaviors in diverse benchmark settings.







\end{document}